\documentclass[letterpaper,journal]{IEEEtran}

\usepackage{amsmath}
\usepackage{amsfonts}
\usepackage{amssymb}

\usepackage{algorithmic}
\usepackage{algorithm}

\usepackage{array}
\usepackage{tabularx}
\usepackage{multicol}
\usepackage{multirow}
\usepackage{booktabs}

\usepackage{textcomp}
\usepackage{stfloats}
\usepackage{url}
\usepackage{verbatim}
\usepackage{graphicx}
\usepackage{float}

\usepackage[numbers]{natbib}

\usepackage{xcolor}
\usepackage[most]{tcolorbox}
\usepackage{lipsum}
\usepackage{afterpage}

\makeatletter
\let\MYcaption\@makecaption
\makeatother

\usepackage[font=footnotesize,labelfont=sf,textfont=sf]{subcaption}

\makeatletter
\let\@makecaption\MYcaption
\makeatother

\usepackage{hyperref}

\begin{document}

\title{MambaCapsule: Towards Transparent Cardiac Disease Diagnosis with Electrocardiography Using Mamba Capsule Network}

\author{
\IEEEauthorblockN{Yinlong Xu, Zitai Kong, Yixuan Wu, Yue Wang, Xiaoqiang Liu, Yingzhou Lu, Jian Wu, Hongxia Xu$^{*}$}

\thanks{$^*$denotes for Corresponding author.}%
\thanks{This research was partially supported by National Natural Science Foundation of China under grants No. T2541004, Zhejiang Key R\&D Program of China under grant No. 2024SSYS0026, Zhejiang Key Laboratory of Medical Imaging Artificial Intelligence, and the Transvascular Implantation Devices Research Institute (TIDRI) under Grant No. KY052025003.}
\thanks{Yinlong Xu and Zitai Kong are with State Key Laboratory of Transvascular Implantation Devices and TIDRI, Zhejiang University, Hangzhou, China~(e-mail: xuyinlong@zju.edu.cn, kongzitai@zju.edu.cn)}%
\thanks{Yixuan Wu and Jian Wu are with School of Public Health, Zhejiang University and Zhejiang Key Laboratory of Medical Imaging Artificial Intelligence, Hangzhu, China~(e-mail: wyx\_chloe@zju.edu.cn, wujian2000@zju.edu.cn)}%
\thanks{Yue Wang is with State Key Laboratory of Transvascular Implantation Devices of The Second Affiliated Hospital and Liangzhu Laboratory, Zhejiang University, Hangzhou, China~(e-mail: ywang2022@zju.edu.cn)}%
\thanks{Xiaoqiang Liu is with Department of Gastroenterology, First Hospital of Quanzhou Affiliated to Fujian Medical University, China~(e-mail: liuxiaoqianghusina@163.com)}%
\thanks{Yingzhou Lu is with School of Medicine, Stanford University, USA~(e-mail: lyz66@stanford.edu)}%
\thanks{Hongxia Xu is with Liangzhu Laboratory and WeDoctor Cloud and TIDRI, Hangzhou, China~(e-mail: Einstein@zju.edu.cn)}}

\markboth{Journal of \LaTeX\ Class Files,~Vol.~14, No.~8, August~2021}%
{Shell \MakeLowercase{\textit{et al.}}: A Sample Article Using IEEEtran.cls for IEEE Journals}

\IEEEpubid{0000--0000/00\$00.00~\copyright~2021 IEEE}

\maketitle

\begin{abstract}
Cardiac arrhythmia, a condition characterized by irregular heartbeats, often serves as an early indication of various heart ailments. With the advent of deep learning, numerous innovative models have been introduced for diagnosing arrhythmias using Electrocardiogram (ECG) signals. However, recent studies  solely focus on the performance of models, neglecting the interpretation of their results. This leads to a considerable lack of transparency, posing a significant risk in the actual diagnostic process. To solve this problem, this paper introduces MambaCapsule, a deep neural networks for ECG arrhythmias classification, which increases the  explainability of the model while enhancing the accuracy.Our model utilizes Mamba for feature extraction and Capsule networks for prediction, providing not only a confidence score but also signal features. Akin to the processing mechanism of human brain, the model learns signal features and their relationship between them by reconstructing ECG signals in the predicted selection. The model evaluation was conducted on MIT-BIH and PTB dataset, following the AAMI standard. MambaCapsule has achieved a total accuracy of 99.54\% and 99.59\% on the test sets respectively. 
These results demonstrate the promising performance of under the standard test protocol.
\end{abstract}


\section{Introduction}
\IEEEPARstart{C}{ardiac} arrhythmia refers to the disruption of the cardiac rhythm caused by abnormal cardiac electrical activity~\citep{nature2022climate}. For the reason that arrhythmia serves as an indicator to advent heart consequence, its accurate detection and classification are imperative for mitigating potential risks and complications. Electrocardiogram~(ECG) is widely used for recording the electrical cardiac activity~\citep{use_of_ECG}. It provides critical information about the cardiac function and health by measuring the cardiac electrical signals during each heartbeat cycle, which makes it a reliable tool for diagnosing heart health.

Recent researches have shown great advantages of deep learning in arrhythmia classification~\citep{ai_in_heart, ecg_survey}, leveraging the information contained in ECG signals. The development of powerful and effective deep learning methods has the potential to significantly improve the accuracy and efficiency of arrhythmia diagnosis, and ultimately leads to better patient outcomes.  

Although the numerous previous models have made substantial progress in arrhythmia classification, most of them solely focus on the performance  neglecting the interpretation of their results, which leads to a considerable lack of transparency and poses a low reliability in the actual diagnostic process~\citep{cnn2018}.

To alleviate this limitation, we proposed a encoder-decoder based neural network called \textbf{MambaCapsule} in this paper. Inspired by the mechanism of brain processing~\citep{luppi2024information, vision_by_brain, cognitive_by_brain}, which tells what features exactly the eyes see and tells the reason why the brain chooses to classify one item to this label not the other one, we novelly designed our model to implement both processes by using the main architecture for prediction and a reconstruct part for explanation. Our evaluation was carried on MIT-BIH dataset~\citep{mit-dataset} and PTB~\citep{PTB} dataset, reached 99.54\% and 99.59\% accuracy. By applying a reconstruct network, we could intuitively know what the model sees and why it makes that decision.

The main contributions of this paper are as follows:
\begin{itemize}
\itemsep=0pt
\item[$\bullet$] We introduced a novel encoder-decoder based network architecture based on \textbf{Mamba}~\citep{Mamba} and \textbf{Capsule networks}~\citep{capsule}, changing the form of the output from the probability sequence to capsule vector sequence, which contains both the classification and feature attribute.
\item[$\bullet$] We applied a downstream network to reconstruct the ECG signal. The reconstructed signal can be used to explain what our model learns and the reason why our model makes its prediction.
\item[$\bullet$] We applied a fusion states SSM architecture in process of feature extraction. The evaluation indicators show the feature extracted has considerable advantages on the MIT-BIH and PTB datasets.
\end{itemize}  
\IEEEpubidadjcol

\section{Related work}
\subsection{Mamba Neural Network}
Mamba has made excellent performance in numerous traditional fields since it was proposed by \cite{Mamba}. Different from traditional and the most popular structure Transformer, the advantages of Mamba are mainly reflected in the computational efficiency and the ability to deal with long sequences. For the reason above, many researches have been carried to replace traditional components with Mamba\cite{yue2024medmamba, ma2024u}. Wang et al. \cite{graph_mamba} proposed Graph-Mamba, which improved the remote context modeling ability of attention mechanism by integrating Mamba block and node selection mechanism, and solved the problems of high computing cost and limited scalability of attention mechanism. 
Rimon et al. \cite{reinforce_mamba} proposes a new model-based meta-reinforcement learning method, achieving higher returns and better sample efficiency (up to 15x) with little need for hyper-parameter tuning.
Qiao et al. \cite{vl_mamba} proposes a multi-modal large language model VL-Mamba based on state space model, demonstrating competitive performance on a variety of multi-modal benchmark tasks and the great potential of state-space models in multi-modal learning.

ECG signals are highly time-dependent physiological waveforms, where diagnostic decisions often hinge on sparse yet critical local patterns—such as R-peak anomalies, ST-segment deviations, or premature beats. Conventional state space models (SSMs) employ fixed parameters, limiting their ability to adaptively attend to such input-specific features. In contrast, Mamba incorporates a selective mechanism that dynamically modulates its system parameters (e.g., B, C, and) based on the input sequence. This input-dependent adaptivity enables the model to focus selectively on diagnostically relevant segments while suppressing irrelevant or noisy regions—mimicking the attentive behavior of expert clinicians. Recently, researchers have made several attempts to use Mamba for ECG diagnosis. Najia and Faouzi \cite{najia2025enhanced} integrates CNNs, LSTMs, and Mamba blocks (with SSMs) to improve ECG classification accuracy, addresses the computational complexity of transformer-based methods. Qiang et al. \cite{qiang2024ecgmamba} proposes ECGMamba, an architecture using 1D convolutions for local features and Multi-Path Mamba blocks for global semantics, to address Transformers’ low inference efficiency. Consequently, Mamba’s selective feature extraction is particularly well-suited for ECG analysis, offering both long-range contextual awareness and fine-grained sensitivity to transient pathological signatures.

\subsection{Capsule Neural Network}
Capsule network was first proposed in a 2D handwritten digital image classification task and has excellent performance in the view-invariance~\citep{capsule}. Later on, it was widely applied to point cloud~\citep{zhao20193d}, image processing~\citep{chen2021receptor,vijayakumar2019comparative}, and tabular learning~\citep{chen2023tabcaps}, graph learning~\citep{xinyi2018capsule}. Specifically, Xiang et al. \cite{ms_capsule} proposed a multi-scale capsule network that is more robust and efficient for feature representation in image classification, demonstrating a competitive performance on FashionMNIST
and CIFAR10 datasets. Nguyen et al. \cite{nguyen2019use} introduce
a capsule network that can detect various kinds of attacks. The model uses many fewer parameters than traditional convolutional neural networks with similar performance.

Capsule neural network also has a intrinsic advantage for explaining its results. For instance, Li et al. \cite{li2019capsule} produced a sentiment capsule to identify the informative logic unit and the sentiment based representations in user-item level for rating prediction. By introducing capsules in different layers, Shahroudnejad et al. \cite{shahroudnejad2018improved} demonstrated the possibility of transforming deep networks into transparent networks. This work helps make DNNs' internal processes more understandable and provides the ability to explain their decisions.

\subsection{ECG based Cardiac Disease Diagnosis}\label{ML}
Numerous researches have been carried in the field of ECG signal classification and the most challenging part of it is to extract the features from the signal. The feature extraction methods have experienced from the manual features and automatic features. Manually designed features extracting usually applied by traditional ML methods like Support Vetor Machine~\citep{SVM1} and Random Forest~\citep{rf1}.
Qin et al. \cite{qin2017combining} proposed an effective methods to extract the abnormal beat eigenvectors of low-dimensional ECG and classify them using support vector machine. Majeed and Alkhafaji \cite{majeed2023ecg} proposed a ML model using multi-domain features combined with least-square support vector machine (LS-SVM), which extracts the time and frequency domain features and selects the most relevant ones. Ma et al. \cite{ma2025defying} proposes an orthogonal gradient learning~(OGL) paradigm to solve multi-model forgetting in one-shot neural architecture search~(NAS) supernets, verifying its effectiveness and improving performance of supernets.

With the success of deep learning in vision and NLP fields, its ability has been admitted and has been applied to many other fields, including the ECG signal classification~\citep{chen2022me,chen2024congenital,hu2024personalized,hong2019mina,ribeiro2020automatic,pyakillya2017deep,bian2022identifying,chen2021electrocardio,yan2019fusing}. 
\cite{li2020heartbeat} proposed a 1D residual CNN based on deep residual networks with a 31-layer one-dimensional residual convolutional neural network combined with 2-lead ECG signals for the classification.
Kim et al. \cite{kim2022automatic} proposed a novel model combining residual networks, compressed activation blocks and bidirectional short-duration memory networks. It was evaluated on multiple databases and the results showed that the framework performed well especially for a small number of categories.

Although these methods have their own way to explain the process of the prediction, their lack of accuracy and inadequacy of features extracting hinder their practical applications. Some efforts also focused on the class imbalance problom on ECG datasets. For instance, El-Ghaish and Eldele \cite{ecgtransform} introduced a deep learning model called ECGTransForm ,which combines multi-scale convolution, channel calibration module and two-way Transformer mechanism to improve the ability to capture spatio-temporal features in ECG data and solves class imbalance problem by introducing context-aware loss function. However, these methods were employed through simple data reinforcement using statistical methods, the reinforced data benefits to model performance but has no specific physiological information through the process of the reinforcement.

\section{Methods}

\subsection{Mamba}
State space models (SSMs) \cite{ssm1, ssm2} are defined as 1D-time-invariant systems, which serve as the predecessor of Mamba. These systems aim to map a 1D input sequence $x(t)\in \mathbb{R}^L$ to a 1D output sequence $y(t)\in \mathbb{R}^L$ through a hidden state $h(t)\in \mathbb{R}^N$, which can be represented as the following equations:
\begin{equation}
\begin{aligned}
    &h(t) = Ah(t)+Bx(t)\\
    &y(t)=Ch(t)
\end{aligned}
\end{equation}

where $A\in \mathbb{R}^{N\times N}$, $B\in \mathbb{R}^{N\times L}$, $C\in \mathbb{R}^{L\times N}$. Considering that deep neural networks operate in discrete space, researchers applied a zero-order preservation on the matrix $A$ and $B$, turning the equations to the form as:
\begin{equation}
\begin{aligned}
    &h_t=\overline{A}h_{t-1}+\overline{B}x_t\\
    &y_t=Ch_t
\end{aligned}
\end{equation}

Where $\overline{A}=exp(\bigtriangleup A)$ and $\overline{B}=(\bigtriangleup A)^{-1}(exp(\bigtriangleup A)-I)\cdot (\bigtriangleup B)$ for$\bigtriangleup$ represents the time step in the continuous space. SSMs also code the remote dependency by initializing matrix $A$ as HiPPO~\citep{Hippo} and allow training in the form of a convolution kernel to accelerate the training process. However, the matrix $A$, $B$ and $C$ do not change whatever the input data are, leading to problem of low sensitivity. To solve the problem, Mamba applies a selective method to SSMs, which makes the matrix $B$ and $C$ and $\bigtriangleup$ adapt themself based on the input as follows:
\begin{equation}
\begin{aligned}
    &\overline{B} = s_{B}(\Tilde{x})\\
    &\overline{C} = s_{C}(\Tilde{x})\\
    &\overline{\bigtriangleup} = \tau_{\bigtriangleup}(Parameter+s_{A}(\Tilde{x}))
\end{aligned}
\end{equation}

Where $s_B$, $s_C$ and $s_A$ are linear projections, and $\tau_{\bigtriangleup}$ is a SoftPlus function. The matrix $A$ does not depend on the input directly, it is influenced by the $\overline{\bigtriangleup}$ in the process of the zero-order preservation, making all main parameter inputs adaptable~\citep{Mamba}.
The $\Tilde{x}$ is the form in which the input $x$ is preprocessed by:

\begin{equation}
\begin{aligned}
    &\Tilde{x}=SiLU(Conv_{1D}^{d}(x))&
\end{aligned}
\end{equation}

Where $d$ is the size of the convolution kernel~(normally set to 4). The $k$ is the main hyper-parameter chosen by our research, as it severs as a role in determining how long the model specifically focuses on the signal range. In the following, we use different convolution kernel sizes to represent different Mamba architectures, to be specific, we use $SSM^j$ to represent a Mamba architectures whose convolution kernel size is $j$.

\subsection{Capsule Networks}
The whole network contains three parts: the convolution network, the capsule network and the dynamic routing. The difference between the capsule network and the traditional neural network is that the unit of the neural network is a scalar neuron $x\in \mathbb{R}$, while the unit of the capsule network is a capsule, which is a vector $x\in \mathbb{R}^{N}$.

In a capsule network, there are many capsules in one layer, and each capsule contains two types of information: the given vector's orientation and the vector's length as the probability of the capsule, the next capsule first applies a linear transformation of the previous capsule output as:
\begin{equation}
    \hat{u}_{j|i,L+1}=W_{ij, L}u_{i, L}
\end{equation}

where $u_{i,L}\in \mathbb{R}^{D}$ is the $i$-th output capsule at layer $L$, $W_{ij,L}\in R^{\hat{D}\times D}$ serves as the transformation matrix and the $\hat{u}_{j|i, L+1}\in \mathbb{R}^{\hat{D}}$ represents the contributes from the parent $i$-th capsule to the $j$-th one at layer $L+1$. The model then gets the weight of each contributes through a clustering like routing algorithm. At present, the popular routing algorithms include dynamic routing algorithm and self-routing algorithm. This paper adopts dynamic routing algorithm, which obtains the final weight coefficient through several rounds of iteration. In each iterations, the weight coefficient $c$ is:
\begin{equation}
    c_{ij}=\frac{exp(b_{ij})}{\sum_k exp(b_{ik})}
\end{equation}

In the first iteration the $b_{ij}$ is initialized to 0 and will be updated at the end of each iteration. Then the coefficient is used to calculate the temp capsule:
\begin{equation}
    s_j=\sum_i c_{ij}\hat{u}_{j|i}
\end{equation}

Where $s_j\in \mathbb{R}^{\hat{D}}$ serves as a temp result for an iteration, the capsule network designs a special activation function to transform the length of the capsule into [$0$-$1$] as:
\begin{equation}
    v_j = \frac{||s_j||^2}{1+||s_j||^2}\frac{s_j}{||s_j||} \label{activation}
\end{equation}

Where $v_j\in \mathbb{R}^{\hat{D}}$ serves as the real output capsule in each iteration. The equation \ref{activation} separates the long and short length of the capsules in the dim of $0$ to $1$, making the long capsule close to $1$ and short capsule close to $0$. Then the $b_{ij}$ is updated as:
\begin{equation}
    b_{ij} = b_{ij} + v_j\hat{u}_{j|i} \label{update equation}
\end{equation}

The equation \ref{update equation} serves as an clustering-like algorithm for after each iteration the $b_{ij}$ grows bigger in capsule whose position is  closest to the others. And after the setting iterations have done, the $v_j$ will serve as the output of the capsule networks.

In our implementation, the output layer consists of five primary capsules, each corresponding to one classification. The length of capsule $v_k$ represents the model’s confidence in class $k$ , while its orientation encodes class-specific morphological attributes. This vector-based representation is crucial for the subsequent reconstruction module: by feeding a specific capsule into the decoder, the model can generate a class-conditional ECG waveform that reflects the learned diagnostic features of that category.

During training, the reconstruction network uses the ground-truth capsule (i.e., the capsule corresponding to the true label) to compute the reconstruction loss, ensuring that each capsule learns discriminative and reconstructive features. At inference time, the capsule with the maximum length is selected for both classification and reconstruction. 

\subsection{Model Architecture}

\begin{figure*}[htp]
    \centering
    \includegraphics[width=1.0\linewidth]{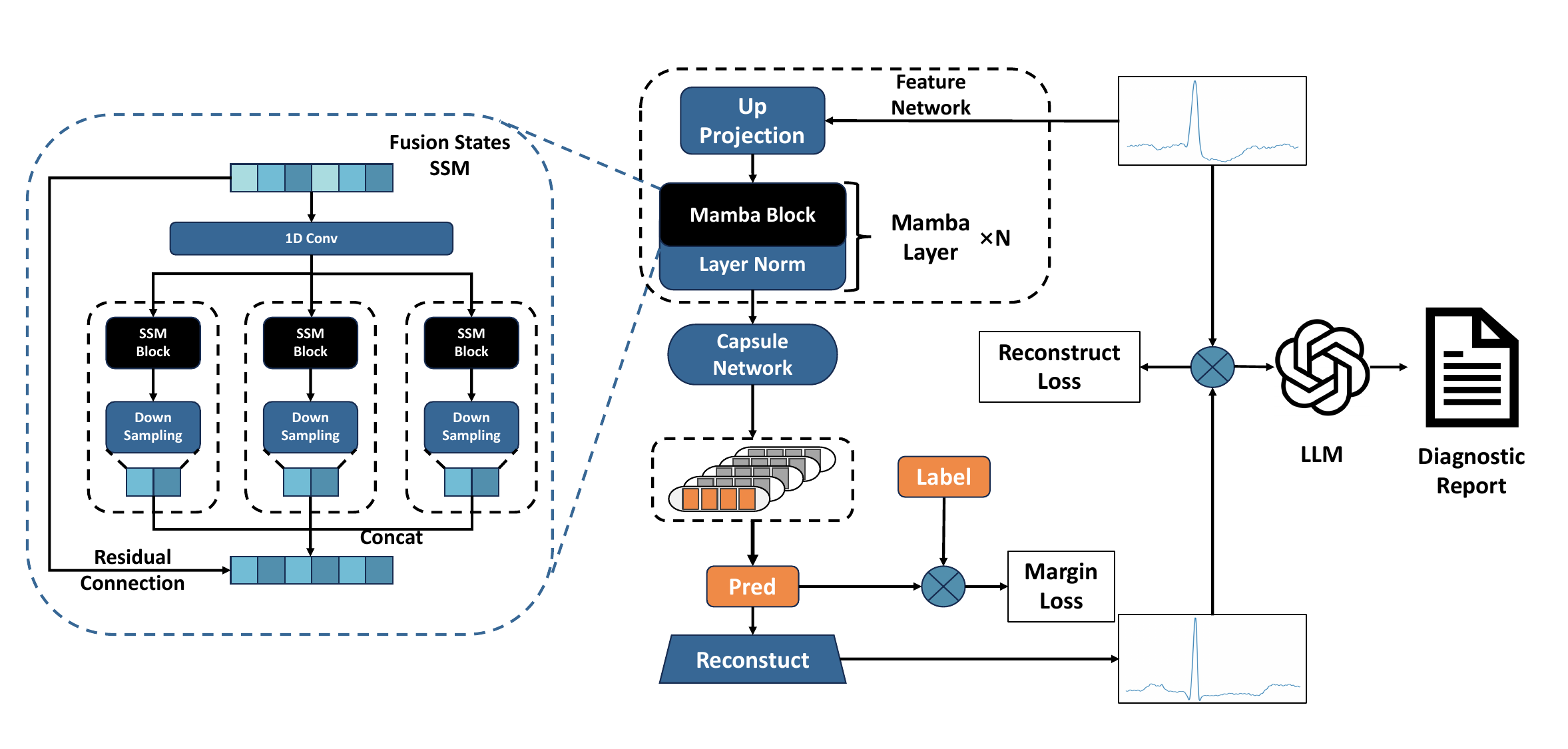}
    \caption{The architecture of MambaCapsule. The model consists of three main parts:the feature network, the capsule network and the reconstruction network. We propose a fusion states SSM to construct backbone of the feature network. The reconstruction loss is calculated as the MSE between the original ECG signal and the signal reconstructed from the Ground-Truth capsule during training.}
    \label{model_structure}
\end{figure*}

In this section, we present a classification model based on Mamba and Capsule networks. The overall framework, as shown in Figure \ref{model_structure}, consists of three main parts: the feature network, the capsule network and the reconstruction network.  Before be fed into the feature network, the input ECG signal
is firstly up-projected to contain scope knowledge not a signal sequence, the input shape changes from $s\in \mathbb{R}^{batch, length}$ to $x_0 \in \mathbb{R}^{batch, length, dim}$, where we set $dim=128$.

Then N~(we set $N=4$) Mamba Layers constitute a pipeline to extract features from the $x_0$ and each Layer contains a Mamba Block and a Layer Norm. Previous work typically apply a single state Mamba as the encoder backbone, however, these fails to extract wider and narrower features. To solve this problem, we propose a fusion states SSM to construct the Mamba Block. For the $i$-th Block, it first applies a 1D-convolution on the upstream $i-1$-th feature sequence $x_i$, which does not change the dimension of the sequence. Then the sequence will be handled by $m$ different SSM Blocks. The different SSM Blocks focus on different scales of time step and have different convolution shape to obtain more abundant and wider features. Then the features will be down sampled and then linked together to add the $x_i$ in the same shape. The whole process can be formulated as:
\begin{equation}
\begin{aligned}
    & \hat{x}_i = Conv_{1D}(x_i)\\
    & u_j = DownSample_j(SSM_j^{d(j)}(\hat{x}_i))\\
    & x_{i+1} = Concat(u_1,\dots,u_m) + x_i\\
\end{aligned}
\end{equation}

Where $u_j \in \mathbb{R}^{batch, length, dim/m}$and the $m$ (4 in our experiments) is the factor of the $dim$ (128 in our experiments)~to assure $Concat(u_1,\dots,u_m) \in \mathbb{R}^{batch, length, dim}$. $SSM^{d(j)}$ denotes for the SSM Block with convolution kernel size of $d(j) \in [4, 8, 12, 16]$ respectively.

The feature network extracts the signal features and sends the features to the capsule network. The capsule network outputs it in the form of capsules in the end. In our experiments, dim of capsule is $D=64$ and layer of it is $L=3$.

The loss of our model consists of two parts, the marginal loss of focusing on model output and the label, and the reconstruct loss focusing on the reconstructed signal and the original signal.

Since our output is in the form of a capsule network, we must first take the geometric length of the output as the probability of the classification label, and then the difference between the set positive label threshold and negative label threshold is made. The final form of margin loss function is as follows:

\begin{equation}
\begin{aligned}
    Loss_{margin} = & \sum_i^k T_{i}max(0, m^{+}-||v_i||)^{2}+\\
    & \lambda(1-T_i)max(0, ||v_i||-m^{-})^{2}
\end{aligned}
\end{equation}

Where $v_i\in \mathbb{R}^{D}$ is the $i$-th output capsule represents one class of the classification, $T_i=1$ if the label of the input is $i$, and $m^{+}=0.9$, $m^{-}=0.1$ at the beginning. $\lambda$ is used to prevent the lengths of vectors from shrinking too much, which is set to $0.5$ in the beginning.

Reconstruct loss needs to choose the label with the highest probability or we choose it for other reasons. Then we put the output into the reconstruct model, which is a 3 fully connection layers model with Relu activation. After that, calculate the MSEloss of the reconstructed signal output of the reconstruct model and the original input signal, the reconstruct loss is formulated as follows:
\begin{equation}
\begin{aligned}
    & \hat{s}_{reconstruct} = Model(v_{choose})\\
    & Loss_{reconstruct}=MSELoss(\hat{s}_{reconstruct}, {s}_{origin})
\end{aligned}
\end{equation}

At the inference stage, we output the reconstructed signal of each capsule and the original signal into a picture, and then all the pictures are given as input to a Multimodal LLM~(MLLM) with parameters $p_{\theta}$, then the MLLM outputs a diagnosis report with a diagnosis reason  according to our requirements:

\begin{equation}
\begin{aligned}
    & Picture_i=Plot(\hat{s}_{reconstruct}^i, s_{origin})\\
    & Report = p_{\theta}(prompt, Concat(\sum_{i}^{k}Picture_i))
\end{aligned}
\end{equation}

Where $\hat{s}_{reconstruct}^i$ denotes the reconstruct signal from the $i-{th}$ capsule, and the $prompt$ is the text prompt, indicating what kind of diagnostic report we want the model to generate.

\section{Model setup and Evaluation}
\subsection{Training Strategy}
\begin{figure}
    \centering
        \includegraphics[width=0.8\linewidth]{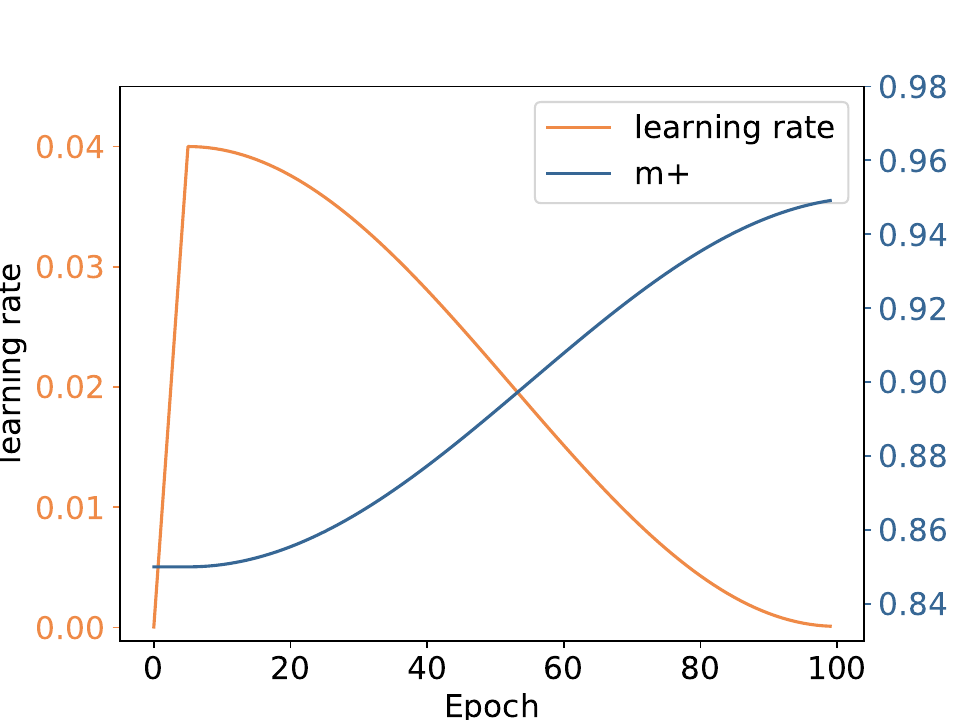}
    \caption{Training set about learning rate and $m^{+}$}
    \label{learning_rate}
\end{figure}

Since the output probability density of the capsule network is not the same as that of the traditional neural classification network, and the bias can directly affect the output, we adjusted bias by cosine schedule during the training process, so that the threshold of bias kept increasing. At the same time, due to the slow training speed of capsule network in the initial stage of training, the cosine annealing algorithm after linear increase is adopted to change the learning rate. The changes of bias and learning rate are shown in the Figure~\ref{learning_rate}.

Our model is trained on a single NVIDIA GeForce RTX 3090 GPU (24 GB). We use the AdamW optimizer with an initial learning rate of 0.001, which is first linearly warmed up for 5 epochs and then decayed using a cosine annealing schedule over a total of 100 training epochs. The batch size is set to 64.

\subsection{Performance Metrics}
To evaluate the model, we applied five performance indexes: accuracy(ACC), sensitivity (SEN), F1-score(F1), precision (PPV) and specificity (SPEC). These indexes are defined as follows:

\begin{equation}
\begin{aligned}
    &ACC=\frac{TP+TN}{TP+TN+FP+FN}\\
    &SEN=\frac{TP}{TP+FN}\\
    &F1=2\times\frac{ACC\times SEN}{ACC+SEN}\\
    &PPV=\frac{TP}{TP+FP}\\
    &SPEC=\frac{TN}{TN+FP}\\
\end{aligned}
\end{equation}

In these formulas, TP represents the number of true positives (correctly labeled positive); TN represents the number of true negatives (correctly labeled negative); FP represents the number of false positives (incorrectly labeled positive); and FN represents the number of false negatives (incorrectly labeled negative).

\begin{table*}[htp]
    \centering
    \caption{The detailed description of adopted datasets.}
    \vspace{5pt}\label{mitbih_dataset}
    \begin{tabular}{c c c c c c c c c c}
        \toprule
        Database & \multicolumn{6}{c}{MIT-BIH Arrhythmia} & \multicolumn{3}{c}{PTB}\\
        \cmidrule(lr){2-7}\cmidrule(lr){8-10}
        Classes & N & S & V & F & Q & Total & Normal & Abnormal & Total\\
        \cmidrule(lr){1-7}\cmidrule(lr){8-10}
        Training & 72470 & 2223 & 5788 & 641 & 6431 & 87553 & 3236 & 8400 & 11636\\
        Testing & 18117 & 556 & 1448 & 162 &1608 & 21891 & 809 & 2100 & 2909\\
        \bottomrule
    \end{tabular}
 \end{table*}
 \subsection{Dataset}
 Our training and testing datasets are preprocessed in the same routine and are randomly splited in a ratio of $0.8$. The detailed description of the dataset is in Table\ref{mitbih_dataset}:
\subsubsection{MIT-BIH Arrhythmia Dataset}
The MIT-BIH arrhythmia dataset \citep{mit-dataset} is widely used for arrhythmia classification in recent researches. The dataset contains 48 half-hour two-channel ECG recordings digitized at 11 bit resolution in 10 mV range and a rate of 360 samples per second per channel. Two cardiologists annotated each record separately (about 110,000 annotations in total). MIT-BIH has annotated 41 categories which can be converted to five-classes classification problems following the ANSL/AAMI EC57$:$ 2012 (The Association for the Advancement of Medical Instrumentation) standard. These distinctive classes are: Normal Sinus Rhythm (N), Supraventricular Premature or Ectopic Beat (S), Ventricular Premature or Ectopic Beat (V), Fusion of Ventricular and Normal Beat (F), Unknown Beats (Q). Some researches were carried on three or four classes due to imbalanced sample size, making the comprehensive evaluation difficult. In this paper, we chose to adapt to all the five classes.

\subsubsection{PTB Diagnostic ECG Database}
The PTB diagnostic ECG database is a popular ECG recording database collected by the National Metrology Institute of Germany\cite{PTB}. The dataset contains 549 records from 290 subjects (aged 17 to 87, mean 57.2; 209 men, mean age 55.5, and 81 women, mean age 61.6; ages were not recorded for 1 female and 14 male subjects). Each subject is represented by one to five records. Each record includes 15 simultaneously measured signals: the conventional 12 leads (i, ii, iii, avr, avl, avf, v1, v2, v3, v4, v5, v6) together with the 3 Frank lead ECGs (vx, vy, vz). Each signal is digitized at 1000 samples per second, with 16 bit resolution over a range of ± 16.384 mV. The PTB dataset consists of two main classes, i.e., (1) Normal (N), which represents normal ECG recordings, and (2) the Myocardial Infarction (M), which represents ECG recordings that exhibit signs of myocardial infarction, indicating the presence of a heart attack.

\subsubsection{Data Preprocess}
The raw ECG recordings from the MIT-BIH and PTB Database are often contaminated by various types of noise, including power-line interference (typically at 50/60 Hz), electromyographic (EMG) artifacts, and baseline wander. To mitigate these disturbances while preserving diagnostic morphological features, we employed a denoising strategy based on the Discrete Wavelet Transform (DWT) with the Daubechies 6 (db6) wavelet~\cite{sun2023dynamic}. The db6 wavelet was selected for its balance between smoothness and compact support, which aligns well with the transient characteristics of ECG waveforms (e.g., sharp QRS complexes and smooth P/T waves).

Specifically, the ECG signal was decomposed into multiple levels (levels $6-8$) using DWT. Coefficients corresponding to high-frequency noise (levels $1–2$) and very low-frequency baseline drift (the approximation at the highest level) were zeroed, while the intermediate levels containing clinically relevant components (P wave, QRS complex, T wave) were retained. The denoised signal was then reconstructed via the inverse DWT. Following denoising, single-beat segments were extracted by centering a fixed-length window (187 samples at 360 Hz) around each annotated R-peak, and each beat was normalized to zero mean and unit variance. This pipeline yielded clean, standardized single-heartbeat ECG samples suitable for downstream modeling.

\section{Experimental Results}
\subsection{Performance Evaluation}

Many previous work focused on only four or less classes performance, aiming to reduce the perplexity of the problem or concentrate on specific arrhythmia types. However, it leads to great difficulties to compare the ability and universality of different models. In our work, we operate our evaluation on five full classes.
\begin{table}[h]
    \centering
    \caption{Confusion matrix of model prediction on MIT-BIT dataset}
    \label{confusion_matrix}
    \begin{tabular}
    {c|c c c c c}
         \toprule
          & N & S & V & F & Q\\
         \midrule
         N & 18092 & 11 & 14 & 0 & 1\\
         S & 153 & 401 & 1 & 0 & 1\\
         V & 3 & 7 & 1422 & 11 & 5\\
         F & 29 & 0 & 11 & 122 & 0\\
         Q & 1 & 0 & 6 & 0 & 1601\\
         \bottomrule
    \end{tabular}
\end{table}

\begin{table}[h]
    \centering
    \caption{The detailed results for each class on MIT-BIH dataset.}
    \label{score_of_each_classes}
    \begin{tabular}{c|c c c c c|c}
         \toprule
         \multirow{2}{*}{Metrics} &
         \multicolumn{5}{c|}{Per-class performance} &
         \multirow{2}{*}{Macro-Avg}\\
         \cmidrule(lr){2-6}
          & N & S & V & F & Q & \\
          \cmidrule(lr){1-7}
          Accuracy & 99.03 & 99.21 & 99.74 & 99.77 & 99.94 & 99.54\\
          Sensitive & 99.86 & 72.12 & 98.20 & 75.31 & 99.56 & 89.01\\
          F1-score & 99.44 & 83.52 & 98.96 & 85.29 & 99.75 & 93.50\\
          Precision & 98.98 & 95.70 & 97.80 & 91.73 & 99.56 &96.76\\
          Specificity & 95.07 & 99.92 & 99.84 & 99.95 & 99.97 &
          98.95\\
          \bottomrule
    \end{tabular}
\end{table}

Table \ref{confusion_matrix} shows the confusion matrix of our best model prediction. And the detailed evaluations on each class of the MIT-BIH dataset are presented in Table\ref{score_of_each_classes}. The result shows our model has excellent ability of discrimination, especially in class N, V and Q the Sensitive, Precision and Specificity reached almost 100\%, which can be attribute the large training sample size. The performance of Supraventricular Premature or
Ectopic Beat (S) is not so outstanding, which can be the result of lack of training samples and the Fusion of Ven-tricular. The average performance of Normal Beat (F) can be the reason its original features resemble to the Normal class (N).

\begin{table}[h]
    \centering
    \caption{Confusion matrix of model prediction on PTB dataset}
    \label{confusion_matrix_PTB}
    \begin{tabular}{c|c c}
         \toprule
          & N & M\\
         \midrule
         N & 800 & 9\\
         M & 4 & 2096\\
         \bottomrule
    \end{tabular}
\end{table}

\begin{table}[h]
    \centering
    \caption{The detailed results for each class on PTB dataset.}
    \label{score_of_each_classes_PTB}
    \begin{tabular}{c|c c|c}
         \toprule
         \multirow{2}{*}{Metrics} &
         \multicolumn{2}{c|}{Per-class performance} &
         \multirow{2}{*}{Macro-Avg}\\
         \cmidrule(lr){2-3}
          & N & M & \\
          \cmidrule(lr){1-4}
          Accuracy & 99.59 & 99.59 & 99.59\\
          Sensitive & 99.24 & 99.72 & 99.48\\
          F1-score & 98.89 & 99.86 & 99.37\\
          Precision & 99.63 & 99.57 & 99.60\\
          Specificity & 99.86 & 98.89 & 99.37\\
          \bottomrule
    \end{tabular}
\end{table}

Table \ref{confusion_matrix_PTB} and Table \ref{score_of_each_classes_PTB} also show our model great performance on PTB dataset. The confusion matrix indicates that our model predicts nearly all of positive and Negative samples, with an macro-avg accuracy of $99.59\%$.
\begin{table*}[h]
    \centering
    \caption{Comparison of models with different architectures on MIT-BIH dataset}
    \label{compare_different_models}
    \begin{tabular}{c c c c c}
         \toprule
         Baseline & Methods & Accuracy & F1-score\\
        \midrule
         Ours & Mamba + Capsule Networks & \textbf{99.54} & \underline{93.50}\\
         El-Ghaish and Eldele \cite{ecgtransform} & MSC + CRM + BiTrans + CAL & \underline{99.35} & \textbf{94.26}\\
         Xia et al. \cite{xia2023transformer} & CNN + DAE + Transformer & 97.66 & -\\
         Nurmaini et al. \cite{nurmaini2020deep} &DAE + DNN &99.34&91.44\\
         Jin et al. \cite{jin2021novel}&DLA + CLSTM & 88.76 & 80.54\\
         Kim et al. \cite{kim2022automatic}&ResNet+ SE block + biLSTM&99.20&91.69\\
         Hammad et al. \cite{hammad2020multitier}&ResNet + LSTM + GA&98.00&89.70\\
         Sellami and Hwang\cite{sellami2019robust}&CNN&95.33&80.08\\
         Pokaprakarn et al. \cite{pokaprakarn2021sequence} & Seq2Seq + CRNN &97.60 &89.00\\
        \bottomrule
    \end{tabular}
\end{table*}

\begin{table*}[h]
    \centering
    \caption{Comparison of performance of labels on MIT-BIH dataset}
    \label{Comparison_of_models}
    \resizebox{\textwidth}{!}{
    \begin{tabular}{cc|ccc|ccc|ccc|ccc|ccc}
         \toprule
         \multirow{2}{*}{Baseline} & \multirow{2}{*}{Acc} & \multicolumn{3}{c}{N} & \multicolumn{3}{c}{S} & \multicolumn{3}{c}{V} & \multicolumn{3}{c}{F} & \multicolumn{3}{c}{Q}\\

         \cmidrule(lr){3-5}\cmidrule(lr){6-8}\cmidrule(lr){9-11}\cmidrule(lr){12-14}\cmidrule(lr){15-17}\\[-2ex]
         
          & & SEN & PPV & SPEC & SEN & PPV & SPEC & SEN & PPV & SPEC & SEN & PPV & SPEC & SEN & PPV & SPEC \\

          \cmidrule(lr){1-17}

          Ours & \textbf{99.54} & 
          \textbf{99.86} & 98.98 & 95.07 &
          72.12 & \textbf{95.70} & \textbf{99.92} &
          \textbf{98.20} & \textbf{97.80} & \textbf{99.84} &
          75.31 & \underline{91.73} & \textbf{99.95} &
          \textbf{99.56} & \textbf{99.56} & \underline{99.97}\\

          El-Ghaish and Eldele \cite{ecgtransform}& \underline{99.35}&
          \underline{99.46}& \textbf{99.36}& -&
          86.91& 91.67& -&
          \underline{97.65}& 95.74& - &
          82.68& 91.30& -&
          \underline{99.06}& \underline{99.30}& -\\

          Xia et al. \cite{xia2023transformer}&97.66 
          &97.35 &96.47 &71.09 
          &70.26 &82.90 &99.44 
          &73.92 &71.67 &96.42 
          &13.40 &30.41 &99.08
          &- &- &\textbf{99.98}\\
          
        Hammad et al. \cite{hammad2020multitier} &98.00&
        98.40&99.40&95.30&
        90.00&79.10&99.20&
        95.10&91.80&99.30&
        \underline{88.80}&91.10&99.70&
        33.30&25.00&99.90\\
        
          Acharya et al. \cite{acharya2017deep} & 97.37&
          91.64&85.17&\underline{96.01}&
          89.04&\underline{94.76}&98.77&
          94.07&95.05&98.74&
          \textbf{95.21}&\textbf{94.69}&98.67&
          97.39&98.40&99.61\\
          
          Sellami and Hwang \cite{sellami2019robust}& 95.33 & 
          88.51 & 98.80 & 91.30 & 
          82.04 & 30.44 & 92.80 & 
          92.05 & 72.13 & 97.54 & 
          68.30 & 26.58 & 98.52 & 
          14.29 & 50.00 & 95.33\\

          Marinho et al. \cite{marinho2019novel}& 94.30 & 
          99.00 & - & 57.60 & 
          2.70 & - & \underline{99.80} & 
          87.80 & - & 99.40 & 
          49.50 & - & 99.60 & 
          - & - & 94.30\\

          Li et al. \cite{li2019automated} & 91.44 & 
          91.81 & 98.92 & 91.65 &
          \textbf{95.15} & 90.11 & 93.62 & 
          89.05 & 35.41 & 99.21 &
          32.22 & 20.36 & 98.93 & 
          - & - & 91.44\\

          Chen et al. \cite{chen2020multi}& 96.77 & 
          99.15 & 97.81 & 81.96 & 
          63.90 & 76.10 & 99.23 & 
          88.84 & \underline{96.78} & \underline{99.80} & 
          47.68 & 54.25 & 99.68 & 
          - & - & -\\

          Wang et al. \cite{wang2020high}& 96.72 & 
          95.05 & 98.35 & 87.07 & 
          90.25 & 43.50 & 95.50 & 
          84.13 & 89.52 & 99.32 & 
          1.29 & 5.32 & \underline{99.82} & 
          - & - & -\\

          Shi et al. \cite{shi2019hierarchical}& 92.07 & 
          92.13 & 99.45 & 95.64 & 
          \underline{91.67} & 46.22 & 95.51 & 
          95.12 & 88.09 & 99.04 & 
          61.60 & 15.16 & 97.12 & 
          - & - & -\\

          Dias et al. \cite{dias2021arrhythmia} & 90.29 & 
          79.64 & \underline{99.51} & \textbf{98.09} & 
          91.32 & 40.34 & 94.81 & 
          87.26 & 93.15 & 99.55 & 
          81.14 & 4.50 & 86.24 & 
          - & - & -\\
          \bottomrule
    \end{tabular}}
\end{table*}

\begin{figure}[ht]
    \centering
    \includegraphics[width=0.9\linewidth]{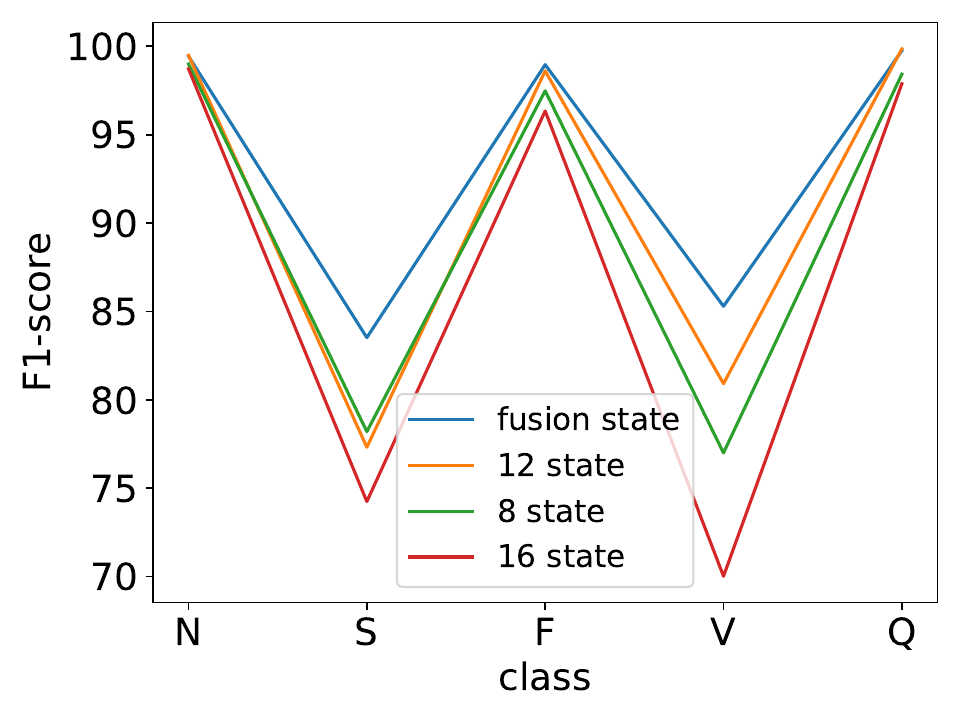}
    \caption{The F1-score of different states.}
    \label{ablation_state}
\end{figure}

\begin{figure}[ht]
    \centering
    \includegraphics[width=0.9\linewidth]{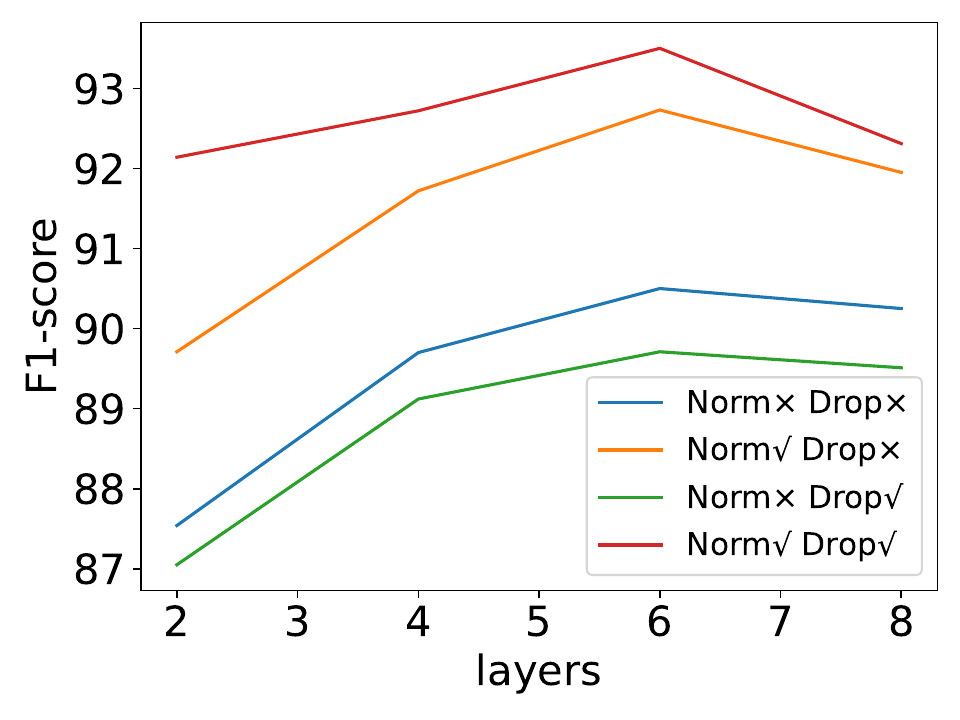}
    \caption{The F1-score of different layers.}
    \label{ablation_layers}
\end{figure}

\begin{figure*}[ht]
    \centering
    \begin{subfigure}{0.325\linewidth}
        \centering
        \includegraphics[width=1.05\linewidth]{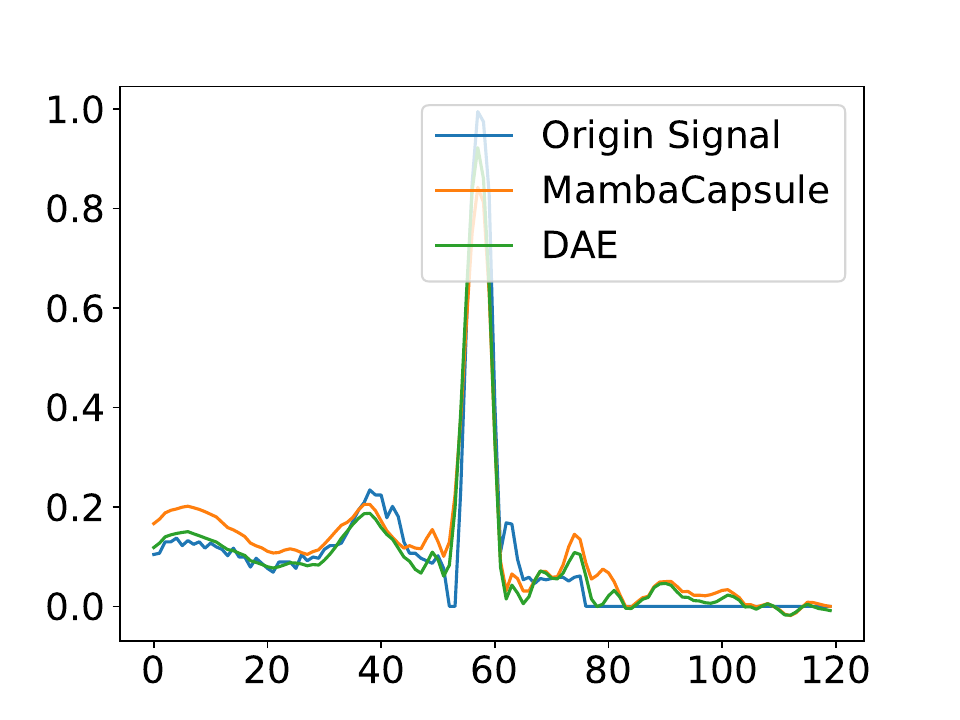}
        \caption{Reconstruction for origin signal.}
        \label{reconstruct_origin}
    \end{subfigure}
    \begin{subfigure}{0.325\linewidth}
        \centering
        \includegraphics[width=1.05\linewidth]{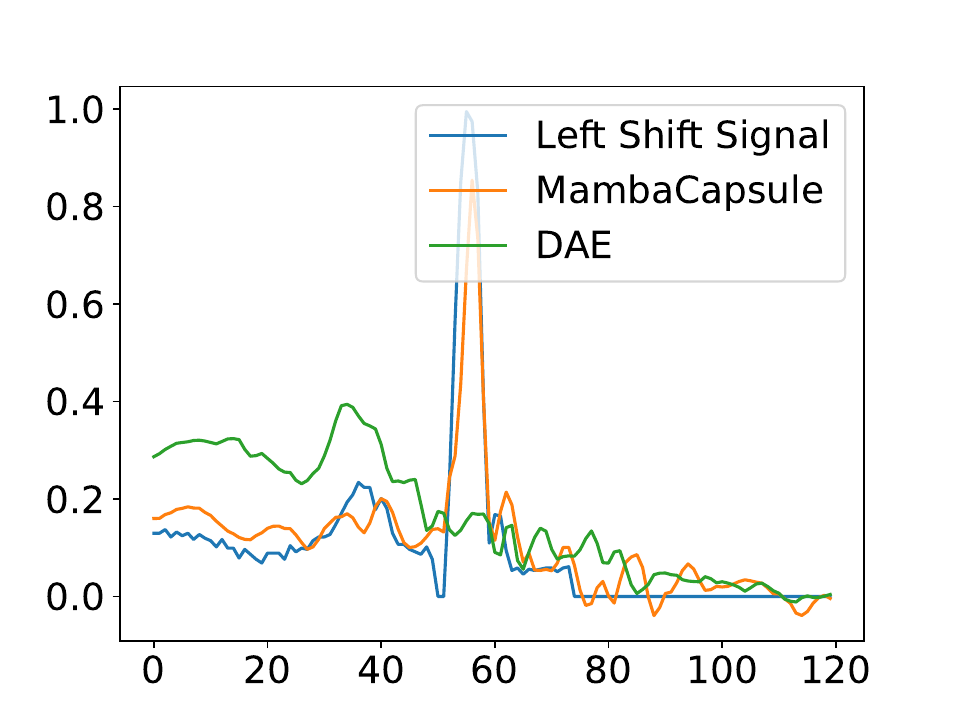}
        \caption{Reconstruction for left shift signal.}
        \label{reconstruct_forward}
    \end{subfigure}
    \begin{subfigure}{0.325\linewidth}
        \centering
        \includegraphics[width=1.05\linewidth]{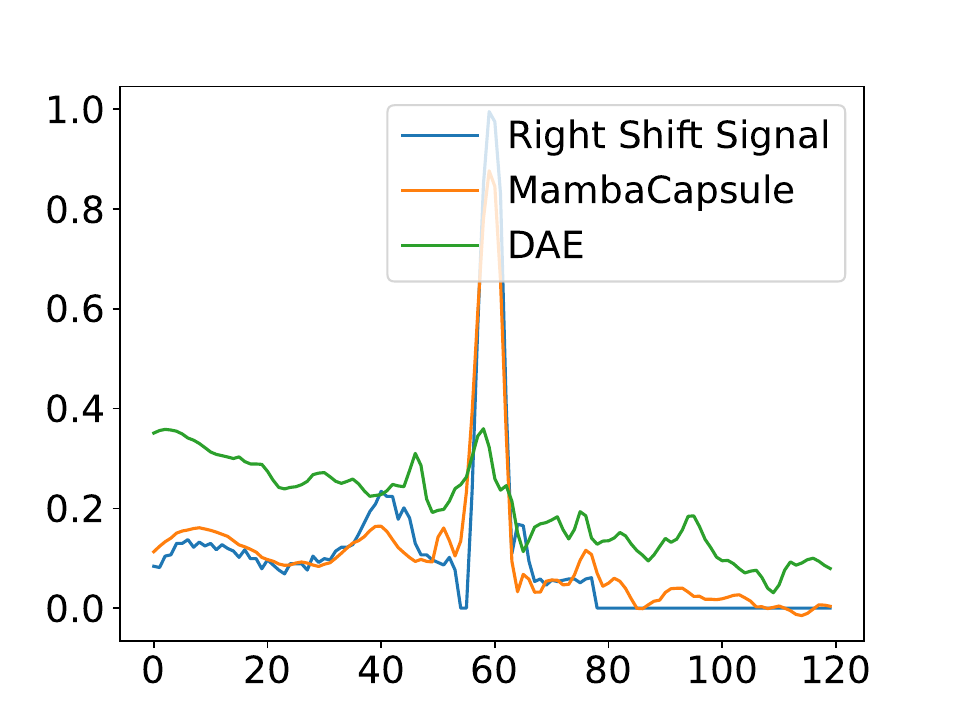}
        \caption{Reconstruction for right shift signal.}
        \label{reconstruct_backward}
    \end{subfigure}
    \caption{Disturbance reconstruction.}
    \label{Disturbance reconstruction}
\end{figure*}

\begin{figure}[ht]
    \centering
    \includegraphics[width=0.9\linewidth]{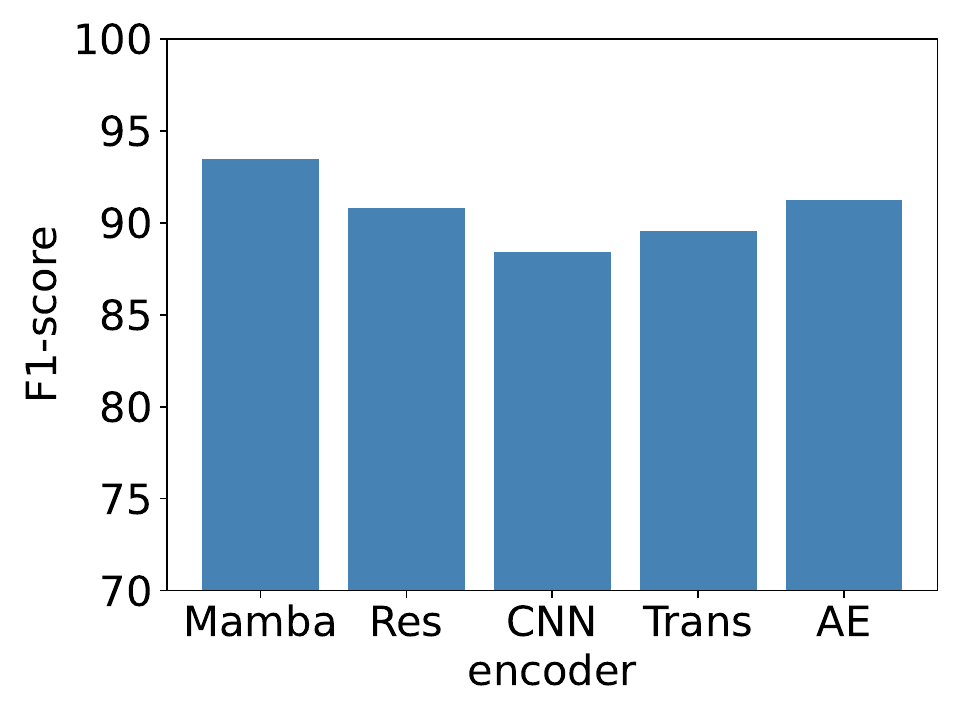}
    \caption{The F1-score with different encoders.}
    \label{ablation_encoders}
\end{figure}

The performance of our model is also compared with several previous studies, which is presented in Table \ref{compare_different_models}, presenting comparative advantages over the other architectures. And the performance of each label can be seen in Table \ref{Comparison_of_models}, which can be observed from the result that our model has considerable advantages compared with others in majority classes.

\subsection{Ablation Study}

The ablation experiment was carried out to test the effectiveness of our model structure. Since \textbf{Mamba} was mainly used as the encoder and capsule networks as the decoder in our model, the ablation experiment was carried out in these two aspects on MIT-BIH dataset.

\subsubsection{Encoder Structure}
This experiment aims to compare the performance of different model structures, including the number of stacked mamba layers, the number of mamba states, the steps per layer, whether to adopt the layernorm and dropout structures, and the use of different models as encoders.

In order to assess the number of mamba states and steps in each layer, we adjusted these parameters while keeping the encoder layers fixed at 4. The experimental results depicted in the Figure~\ref{ablation_state} demonstrate a significant impact of varying numbers of states and steps on the outcomes. Moreover, a layer composed of diverse mamba blocks outperforms a single layer due to its ability to focus on different ECG data features across various time scales. Consequently, this wider temporal coverage achieved by combined blocks leads to enhanced performance. However, if the states are too wide it will miss significant information.

To assess the impact of encoder layers on model performance, we held the parameters of each layer constant and varied the number of encoder layers from 2 to 8 while considering whether to implement layer normalization and dropout. The corresponding results are depicted in the Figure \ref{ablation_layers}. From our analysis of training time and memory usage for a four-layer encoder, it is evident that the metric increases with additional layers; however, model performance does not improve proportionally with increased depth. In fact, the optimal performance is achieved at six layers due to limitations in information capacity exceeding what can be learned from available data as well as increased computational demands associated with larger models.

\begin{figure}[ht]
    \centering
    \includegraphics[width=0.9\linewidth]{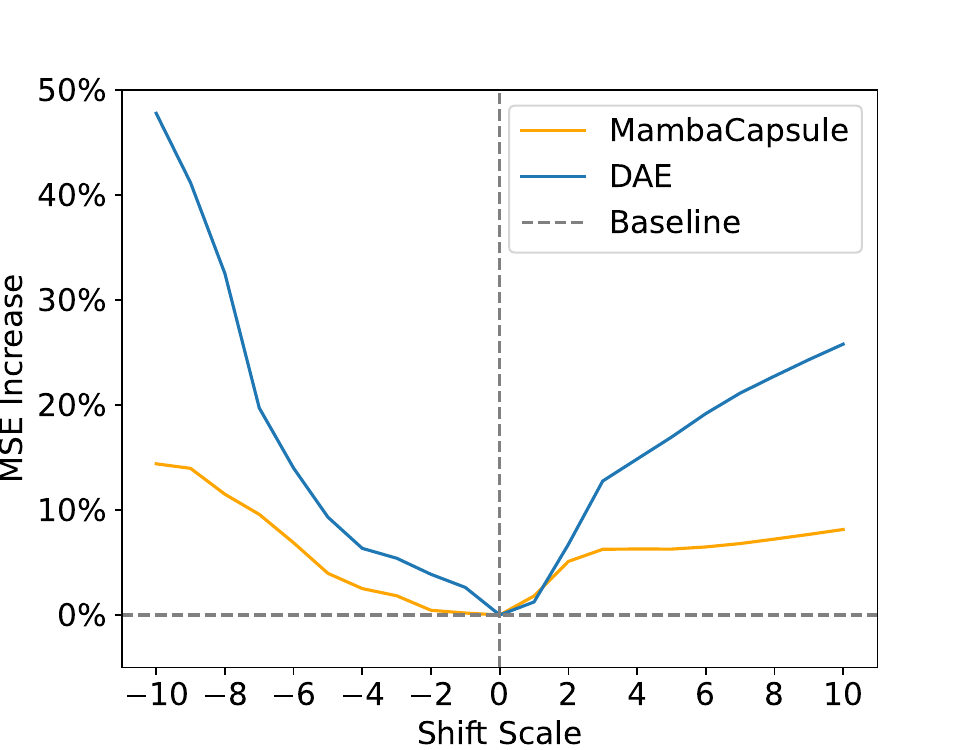}
    \caption{MSE increases with shifted signal.}
    \label{mse of shift}
\end{figure}

\begin{figure}[ht]
    \centering
    \includegraphics[width=0.9\linewidth]{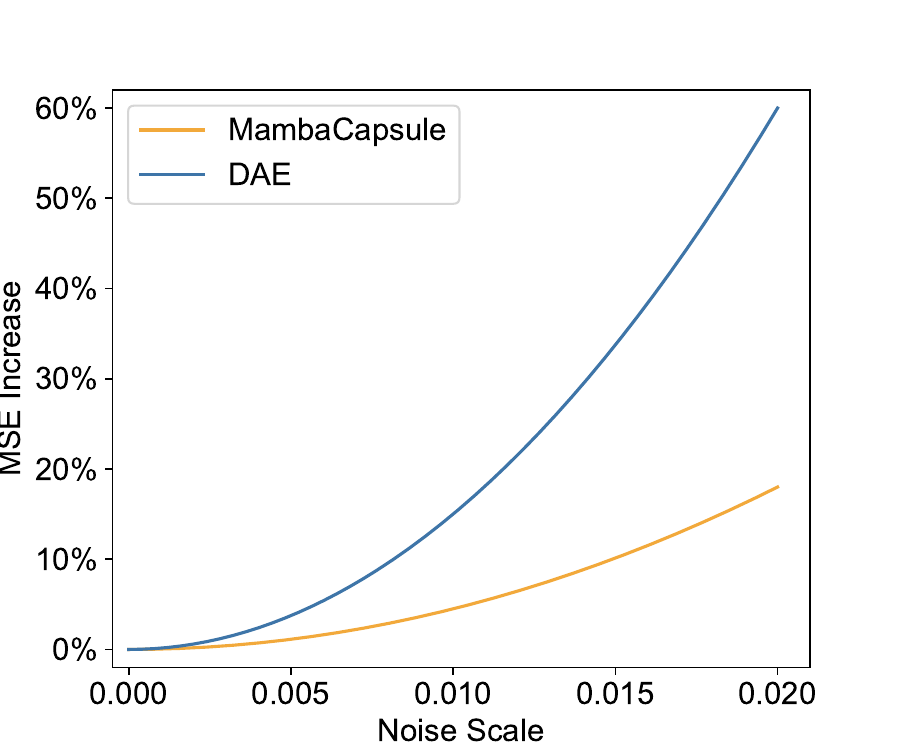}
    \caption{MSE increases with noised signal.}
    \label{mse of noise}
\end{figure}

In order to evaluate the effect of different types of encoder, we have also tried to use resnet, transformer and cnn as encoders. The experimental results are shown in the Figure~\ref{ablation_encoders}. The results show that mamba has the best effect when it is used as encoder.

\begin{figure*}[ht]
    \centering
    \begin{subfigure}{0.32\linewidth}
        \centering
        \includegraphics[width=1.05\linewidth]{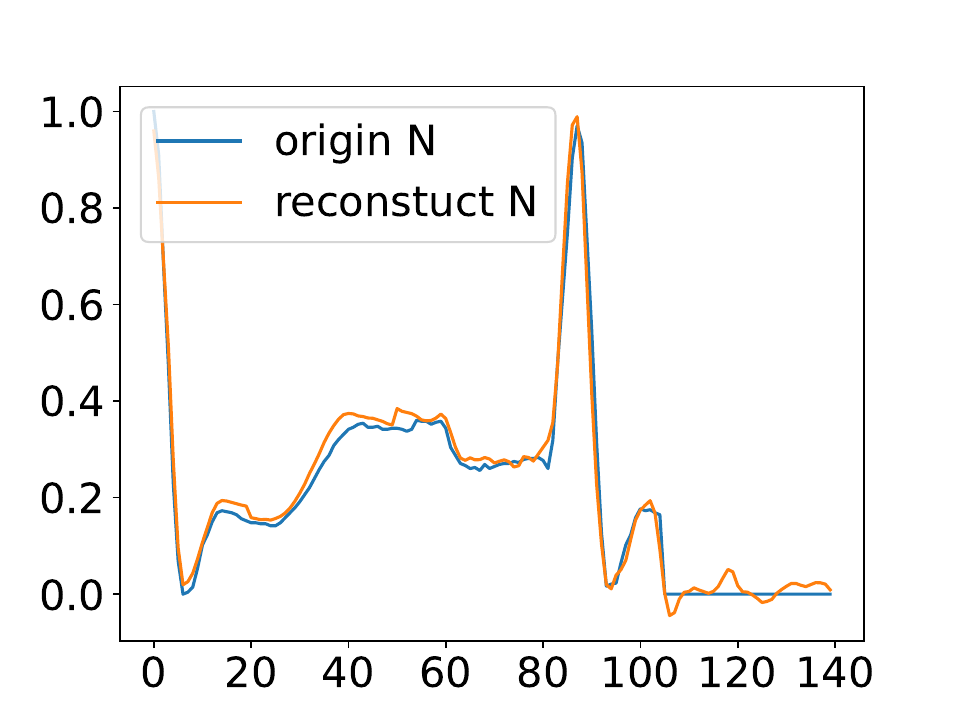}
        \caption{Recontruct signal N from signal N.}
        \label{reconstruct_a}
    \end{subfigure}
    \hfill
    \begin{subfigure}{0.32\linewidth}
        \centering
        \includegraphics[width=1.05\linewidth]{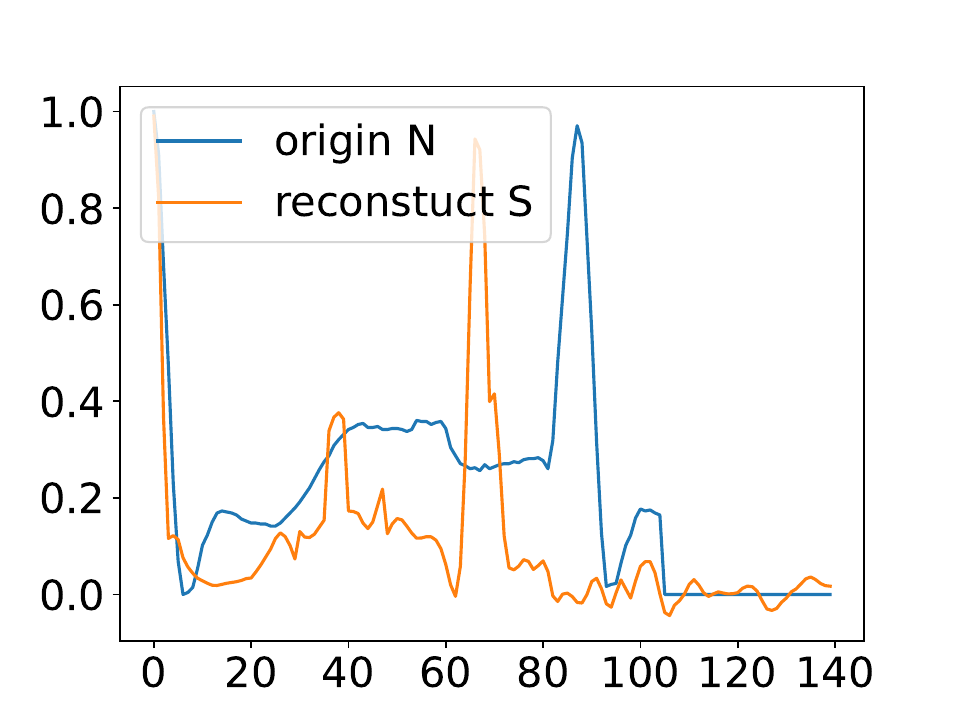}
        \caption{Recontruct signal S from signal N.}
        \label{reconstruct_b}
    \end{subfigure}
    \hfill
    \begin{subfigure}{0.32\linewidth}
        \centering
        \includegraphics[width=1.05\linewidth]{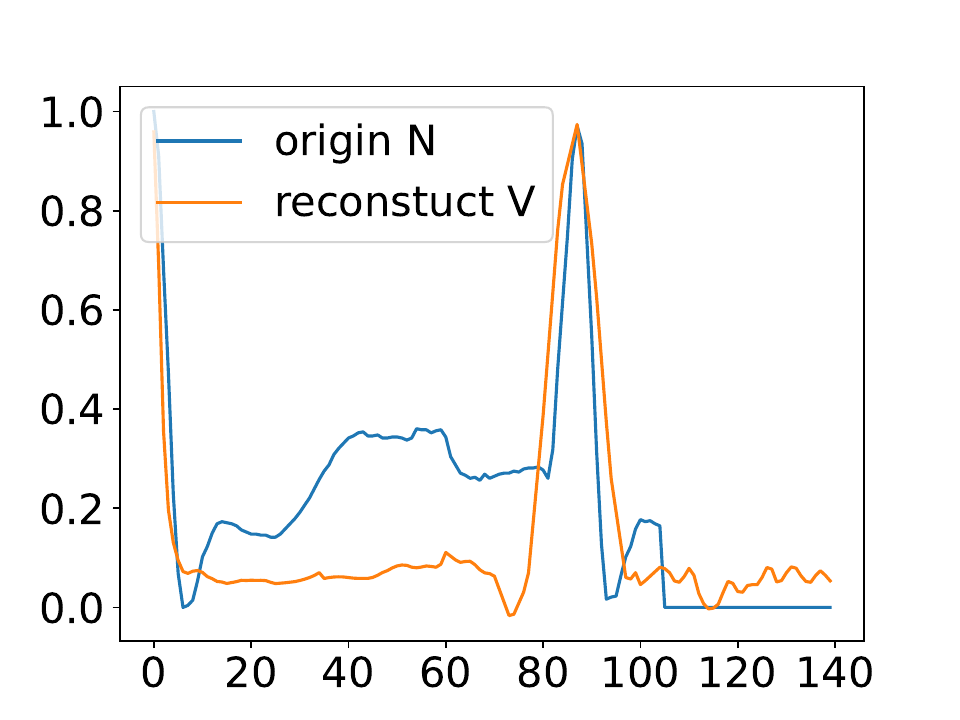}
        \caption{Recontruct signal V from signal N.}
        \label{reconstruct_c}
    \end{subfigure}
    \vspace{0.0cm}
    \begin{subfigure}{0.32\linewidth}
        \centering
        \includegraphics[width=1.05\linewidth]{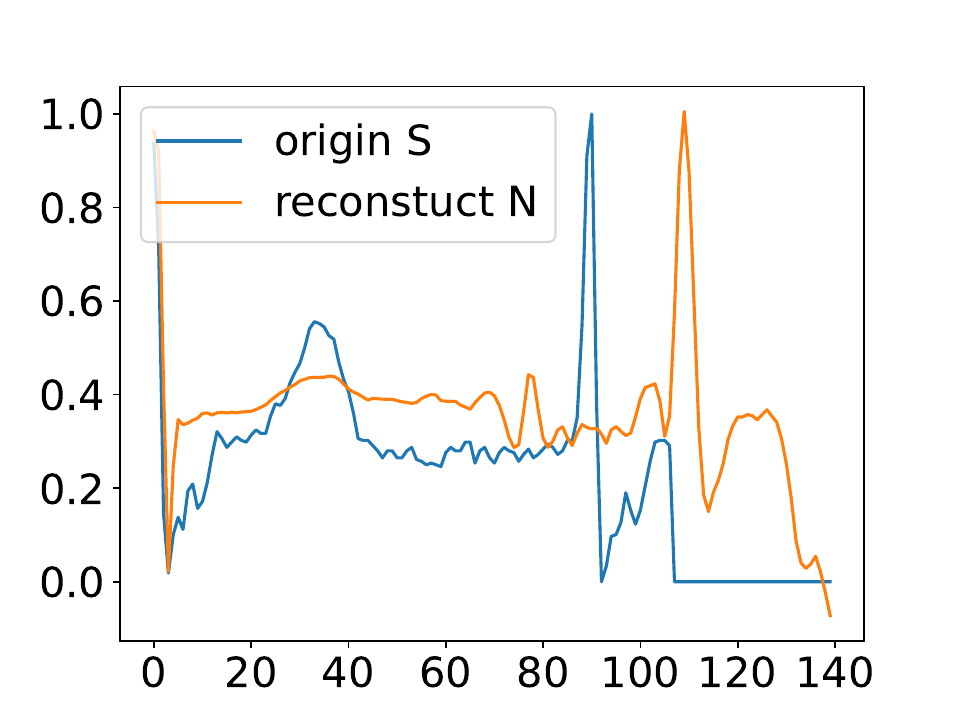}
        \caption{Recontruct signal N from signal S.}
        \label{reconstruct_d}
    \end{subfigure}
    \hfill
    \begin{subfigure}{0.32\linewidth}
        \centering
        \includegraphics[width=1.05\linewidth]{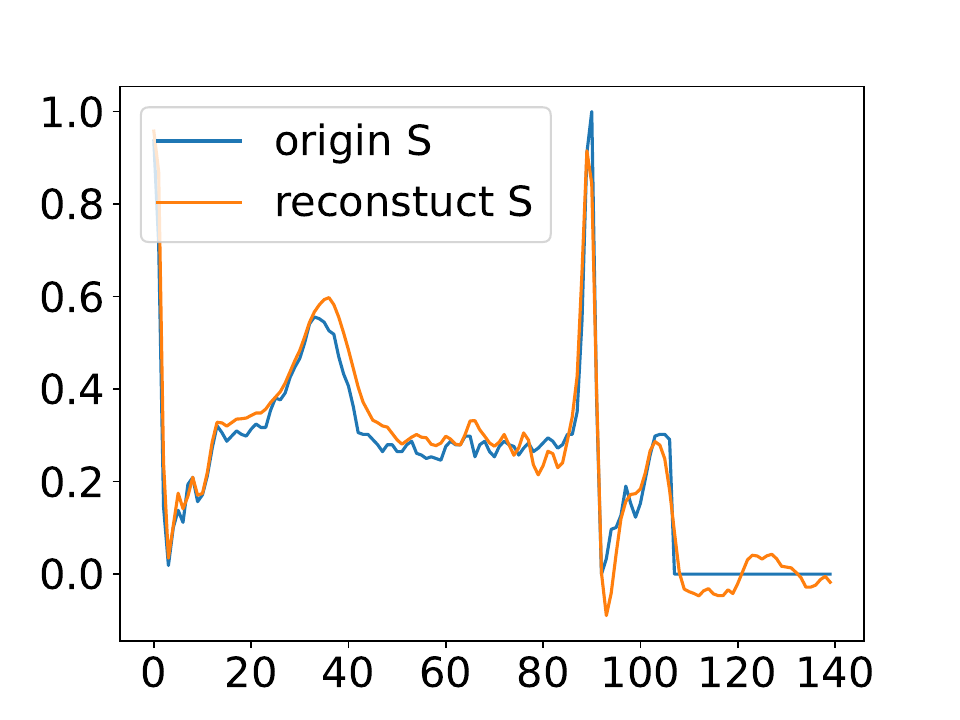}
        \caption{Recontruct signal S from signal S.}
        \label{reconstruct_e}
    \end{subfigure}
    \hfill
    \begin{subfigure}{0.32\linewidth}
        \centering
        \includegraphics[width=1.05\linewidth]{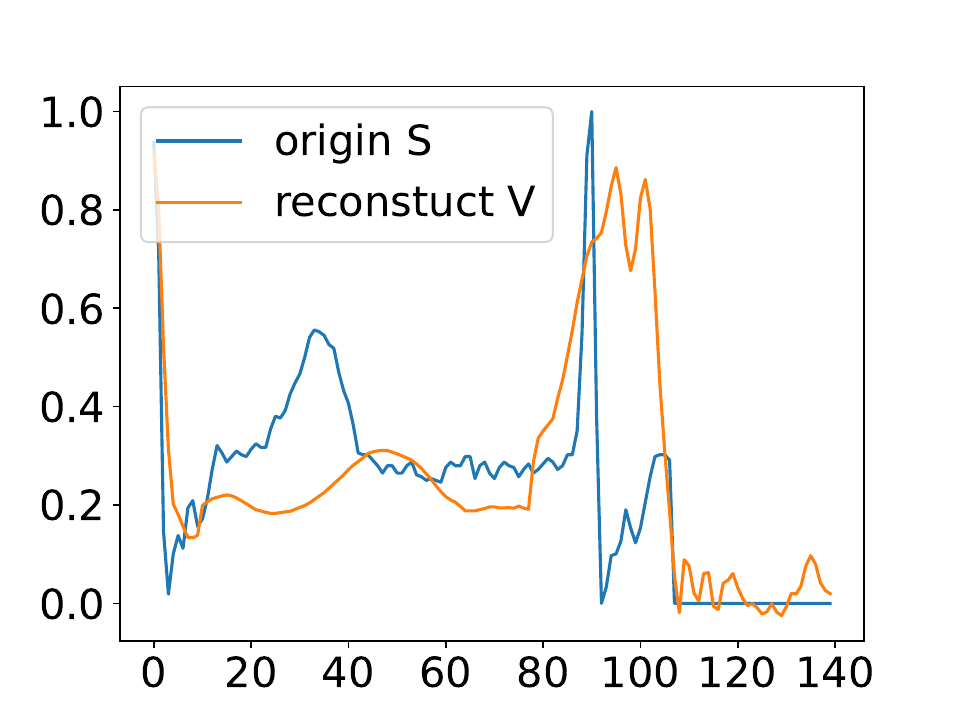}
        \caption{Recontruct signal V from signal S.}
        \label{reconstruct_f}
    \end{subfigure}
    \vspace{0.0cm}
    \begin{subfigure}{0.32\linewidth}
        \centering
        \includegraphics[width=1.05\linewidth]{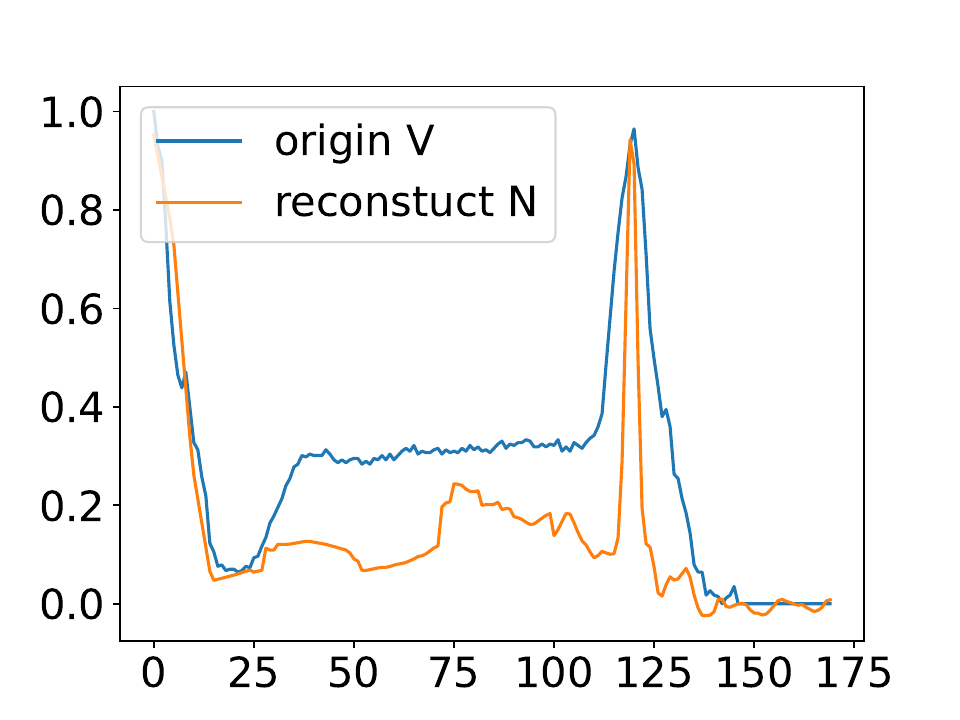}
        \caption{Recontruct signal N from signal V.}
        \label{reconstruct_g}
    \end{subfigure}
    \hfill
    \begin{subfigure}{0.32\linewidth}
        \centering
        \includegraphics[width=1.05\linewidth]{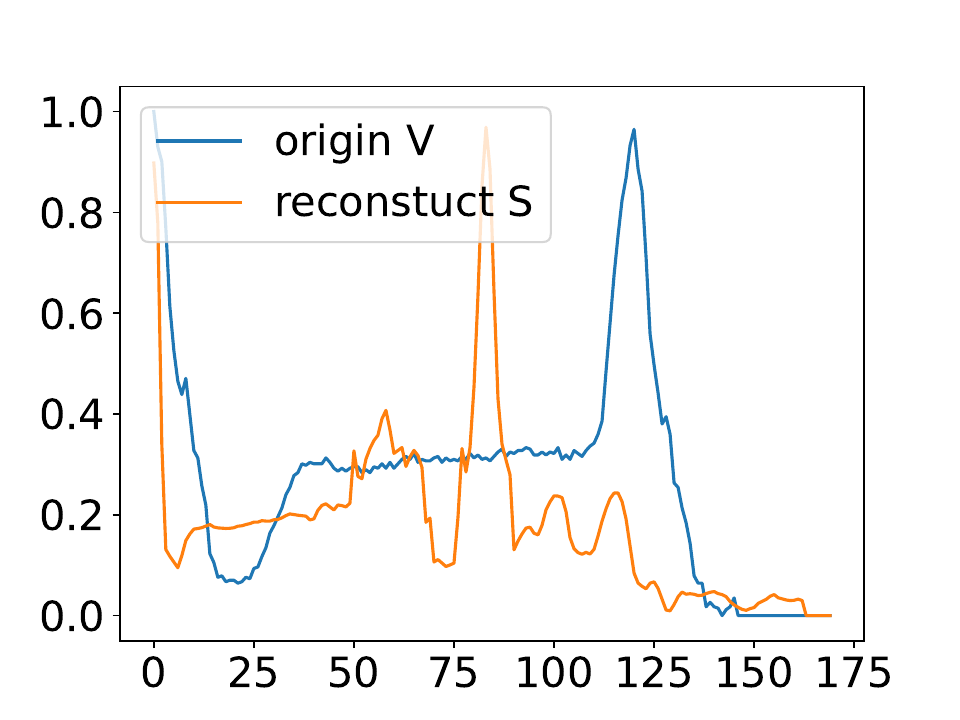}
        \caption{Recontruct signal S from signal V.}
        \label{reconstruct_h}
    \end{subfigure}
    \hfill
    \begin{subfigure}{0.32\linewidth}
        \centering
        \includegraphics[width=1.05\linewidth]{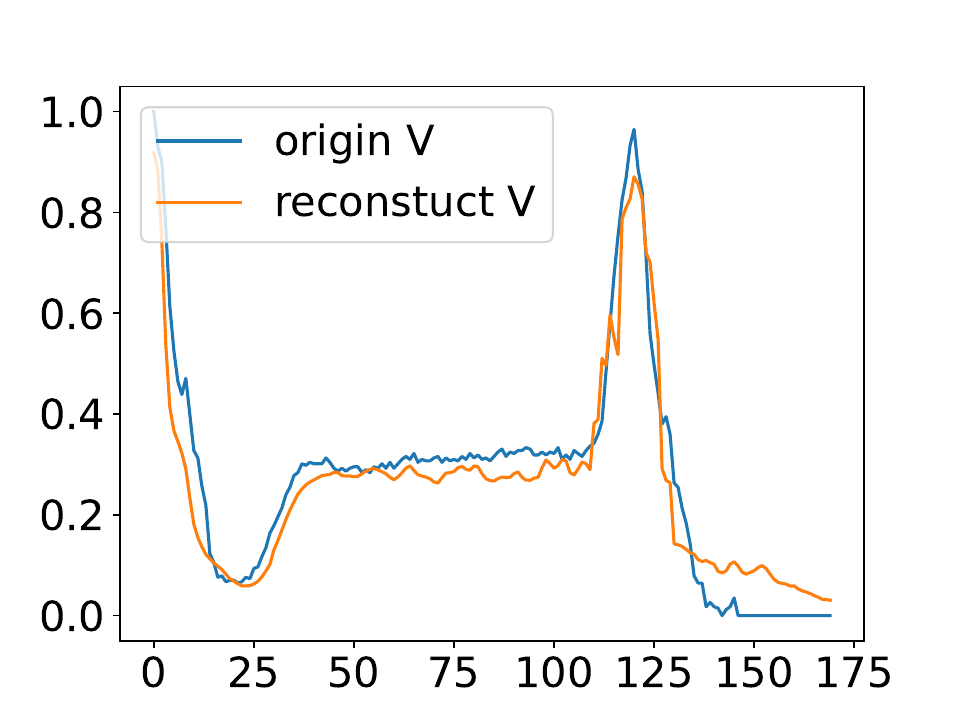}
        \caption{Recontruct signal V from signal V.}
        \label{reconstruct_i}
    \end{subfigure}
    \caption{Different signals reconstruction. Horizontal axis ranges vary across subplots to faithfully represent the intrinsic physiological differences in heartbeat duration for each arrhythmia class.}
    \label{explain reconstruct}
\end{figure*}

\subsubsection{Decoder Structure}
The purpose of this experiment was to compare the changes in the model training and the performance caused by using capsule instead of traditional decoder.

We tried to replace the capsule network with MLP after freezing encoder parameters. Experiments showed that there was no significant difference between the capsule network and MLP under the same encoder with F1-score 93.50 compared to 93.62, but the training time of capsule network was longer than that of MLP. This is determined by the routing structure of the capsule network and the number of parameters. 

\subsection{Robustness Study}

In real-world ECG monitoring, signals are often corrupted by noise and baseline drift due to motion artifacts, electrode contact issues, or environmental interference. To evaluate the model’s reliability under such practical conditions, we conduct a robustness experiment by introducing controlled noise and offsets into the test data, assessing its performance in both classification accuracy and signal reconstruction under realistic degradation scenarios.

\subsubsection{Shift Disturbance Analysis}

When processing the ECG graphs, the human brain not only attends to focus on individual wave peaks but also considers the interrelationship between them. The view invariance of capsule network aligns remarkably well with this characteristic, which is why we selected it as the decoder. In our experiment, we conducted forward and backward disturbance on the input data to observe the disturbance reconstruction of ECG signals using our model and DAE model. The experimental results are depicted in the Figure \ref{Disturbance reconstruction}, where left and right shift represent forward and backward shift respectively. Repeated reconstructions of ECG signals by MambaCapsule exhibit remarkable consistency, indicating that Mambacapsule has learned not only the intrinsic features but also their interrelationships. Conversely, this characteristic is absent in the reconstructed ECG signal produced by DAE model.

We further let the shift value fluctuate in a range to observe the MSE loss changes of the reconstructed signal from the model and the original signal, and the experimental results are shown in Figure \ref{mse of shift}. The results show that with the increase of the offset value of the signal, the MSE loss also increases, and the growth rate of MambaCapsule is significantly slower than that of DAE network, which indicates that MambaCapsule has considerable robustness to offset signal.

\begin{figure*}[ht] 
\centering
\begin{tcolorbox}[
    title={Text and Picture Input for MLLM},
    colback=gray!10,
    colframe=gray!50!black,
    arc=5pt,
    boxrule=1pt,
    width=\textwidth, 
    left=5pt,
    right=5pt,
    top=5pt,
    bottom=5pt,
    before upper={\parindent=0pt}, 
]
    \begin{minipage}[t]{\linewidth}
\setlength{\parindent}{2em}
\noindent You are an expert in ECG signal analysis. I give you three plots, each of which contains an original signal (labeled “origin Signal”) and a reconstructed signal (labeled “reconstruct N/S/V”) for three different classes of reconstruction results. Where N is the Normal Sinus Rhythm, S is the Supraventricular Premature or Ectopic Beat, and V is the Ventricular Premature or Ectopic Beat.\\
Based only on the shape of the waveform in the image, answer:\\
\indent Which class does this original signal most likely belong to? Why not the other two?
Your analysis must be based on the following logic:\\
\indent \textbf{1. Comparison of similarity}: Which type of reconstructed signal is closest to the original signal in terms of overall shape, amplitude, width and timing of key wave groups (such as P wave, QRS wave and T wave)?\\
\indent \textbf{2. Explain the pathological significance of the difference}: For the two classes with the largest difference in reconstruction, describe the typical manifestations of the disease combined with the medical definition, and point out the abnormal features "forced into" the reconstruction signal (e.g., QRS width, P wave loss, ST segment deviation, extra spike, etc.).
    \end{minipage}
    \begin{minipage}[t]{\linewidth}
        \centering
        \begin{tabular}{@{}c@{\hspace{1em}}c@{\hspace{1em}}c@{}}
            \includegraphics[width=0.3\linewidth]{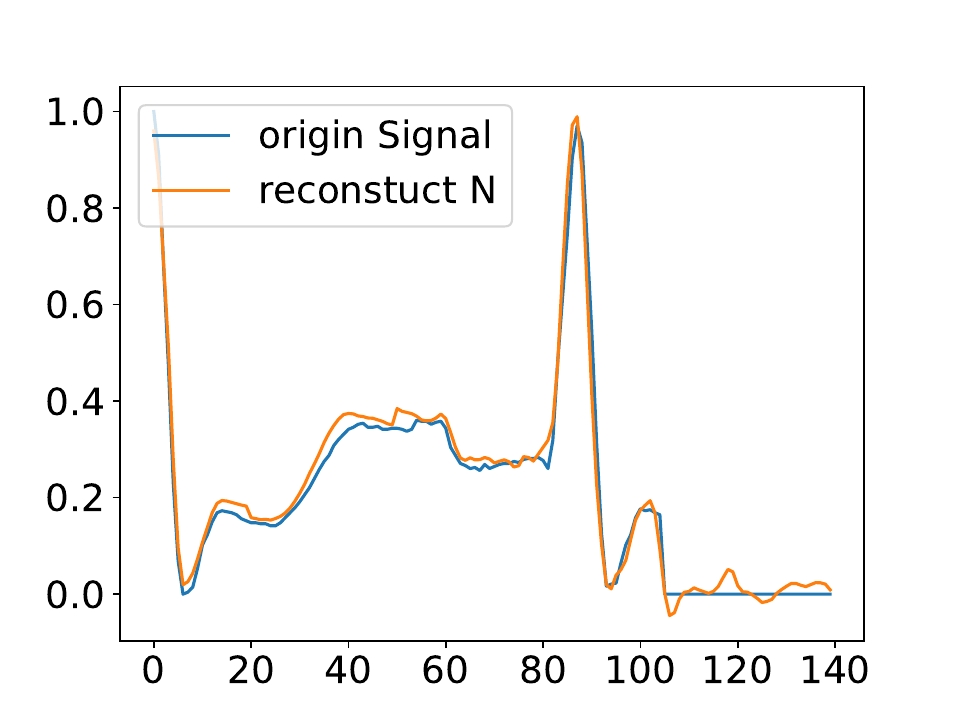} &
            \includegraphics[width=0.3\linewidth]{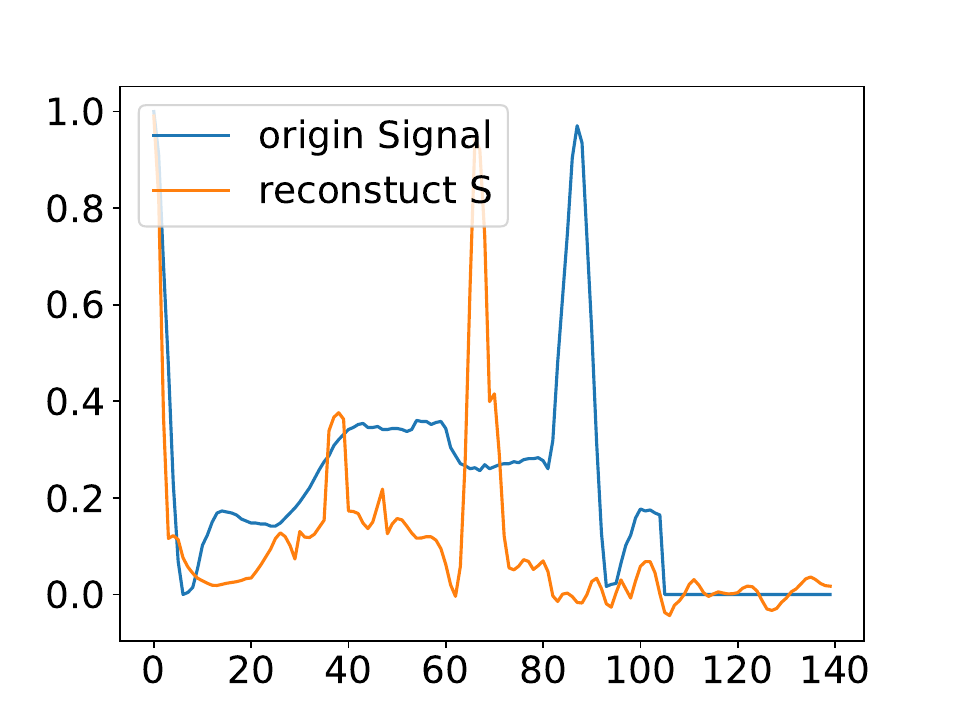} &
            \includegraphics[width=0.3\linewidth]{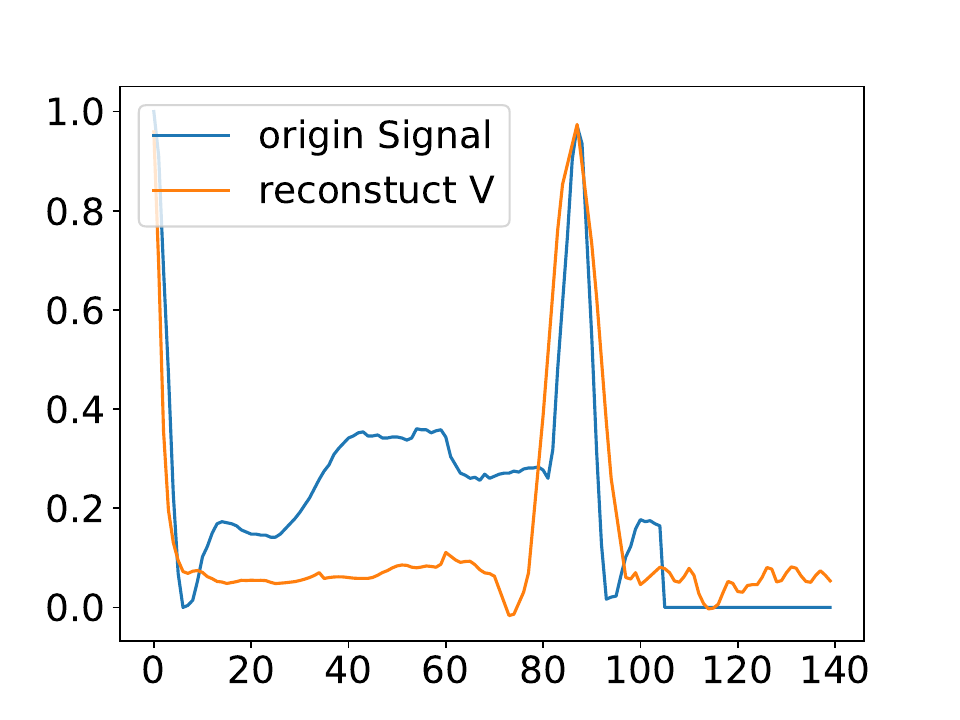} \\
        \end{tabular}
    \end{minipage}
\end{tcolorbox}
\caption{Input format for MLLM.}
\label{MLLM_input}
\end{figure*}

\subsubsection{Noise Disturbance Analysis}

To further evaluate the model’s resilience against common signal degradations in clinical settings, we introduce additive Gaussian noise with varying intensities (controlled by a Noise Scale parameter) into the test ECG signals. Figure \ref{mse of noise} plots the relative increase in MSE loss between the reconstructed signal and the original clean signal as a function of the added noise level.

As shown in Fig. \ref{mse of noise}, both MambaCapsule and the DAE baseline exhibit an increasing MSE with higher noise scales, which is expected. However, the rate at which MSE grows for MambaCapsule is substantially lower than that of the DAE. For instance, at a noise scale of 0.02, the MSE increase for MambaCapsule is approximately 20\%, while the DAE’s MSE increases by nearly 60\%. This indicates that MambaCapsule is more robust to noise perturbations — its feature extraction and reconstruction mechanisms are better at preserving the underlying physiological structure even when the input is corrupted. This enhanced stability under noisy conditions underscores the practical value of our architecture for real-world ECG monitoring systems, where signal quality is often imperfect.

\subsection{Explainable Study}
The interpretability of this paper is unique from other recent papers on ECG diagnosis, mainly through the reconstruction of ECG signals in different ways to explain what knowledge the model has learned on MIT-BIH dataset. 
Considering clinical diagnostic priorities, we focus on P-wave and R-wave reconstruction, as they provide the hallmark features for identifying the AAMI-defined arrhythmia classes (e.g., P-wave for atrial ectopic beats and QRS morphology for ventricular beats). While the T-wave is omitted to optimize computational efficiency, this design choice maintains high diagnostic transparency for rhythm abnormality detection without sacrificing classification accuracy.


\subsubsection{Signal Reconstruction}When human brain categorize an item, it can discern the reasons for its classification and differentiation. Due to the inherent output characteristics of Capsule network, the output not only encompasses the probability density of each category but also encapsulates the distinctive features associated with that particular category. Consequently, by reconstructing ECG signals from unified inputs belonging to different categories, we are able to precisely comprehend why the model deems a certain category more probable. To be specific: given a signal, MambaCapsule will generate five capsules consisting information about different labels respectively then the capsules will be reconstructed to five signals. Although the given signal will only match one capsule, the other capsules will also attempt to extract information that matched the labels represented by the capsules, thus the degree of matching between original signal and reconstruction signal can be serve as an basis for prediction. Meanwhile, the information demonstrated by false capsules can be visualized to explain the reason why the original signal can't be classified in this label.

\afterpage{
\clearpage
\begin{figure*}[ht]
\centering
\begin{tcolorbox}[
    title={Diagnostic report generated by MLLM},
    colback=gray!10,
    colframe=gray!50!black,
    arc=5pt,
    boxrule=1pt,
    width=\textwidth, 
    left=5pt,
    right=5pt,
    top=5pt,
    bottom=5pt,
    before upper={\parindent=0pt},
]
    \begin{minipage}[t]{\linewidth}
\setlength{\parindent}{2em}
\noindent Based on the morphological comparison between the original signal and the reconstructed signal of the three classes in the three images, the original signal most likely belongs to the N class (Normal Sinus Rhythm).\\
The following is a detailed analysis based on the waveform morphology:\\
\textbf{1.Similarity comparison: Why class N (Normal Sinus Rhythm)?}\\
In the first panel (reconstruct N), the orange "reconstruct N" Signal is in high agreement with the blue "origin Signal" on the following key features:\\
\indent 1.1 The overall morphological consistency is high: the contour, fluctuation trend and rhythm of the two waveforms are completely consistent, and there is no obvious phase shift or structural misalignment.\\
\indent 1.2 The P wave is clearly visible: before the QRS complex, there is a gentle, round blunt forward wavelet (P wave), which represents atrial depolarization. This is accurately reconstructed in "reconstruct N", and its amplitude and width match the original signal.\\
\indent 1.3 The QRS complex is narrow and sharp: representing rapid ventricular depolarization with narrow morphology~(0.12 SEC), high amplitude, and steep rise/fall, which is faithfully reproduced in "reconstruct N".\\
\indent 1.4 T wave follows naturally: the shape of T wave after QRS (ventricular repolarization) is round and blunt, and the direction is consistent with the main QRS wave, and they match well.\\
\indent 1.5 Regular rhythm: stable RR interval, no premature or delayed beats, consistent with the characteristics of sinus rhythm.\\
In conclusion, "reconstruct N" is the most faithful morphological restoration of the original signal, indicating that the original signal itself has the typical characteristics of normal sinus rhythm.\\
\noindent \textbf{2. Why not class S (Supraventricular Premature or Ectopic Beat)?}\\
There are significant differences between the reconstructed signal of class S (second Figure) and the original signal, mainly reflected in the following:\\
\indent 2.1 Introduced the "extra premature beat" feature: At approximately x=75, "reconstruct S" presents a preoccurring, morphologically abnormal QRS complex (wider, shorter and morphologically distorted than normal QRS complex) typical of atrial or junctional premature beats.The original signal does not have this premature beat here, and its QRS complex is regular and normal.\\
\indent 2.2 Missing or deformed P-wave:Before premature beats, "reconstruct S" does not show clear P waves, or the relationship between P waves and QRS is abnormal (such as retrograde P waves), which is not consistent with the clear and regular P-QRS-T sequence in the original signal.\\
\indent The rhythm is disrupted: "reconstruct S" introduces irregular rhythms, while the original signal has regular rhythms with no signs of compensatory intermittency or insertional premature beats.\\
Class S represents ectopic beats originating in the atria or atrioventricular node and is characterized by a QRS complex that is usually narrow (unless accompanied by differential transmission) but is often accompanied by missing, distorted, or premature P waves. These "forced introduced" abnormal features do not exist in the original signal, so class S can be excluded.\\
\noindent \textbf{3. Why not class V (Ventricular Premature or Ectopic Beat)?}\\
The reconstructed signal of class V (third Figure) is the most different from the original signal, mainly reflected in the following:\\
\indent 3.1 The QRS complex is forcibly widened and deformed: The QRS complex in "reconstruct V" is significantly widened (>0.12 seconds), thick and blunt, and has an "M" or "W" shape, which are the hallmark features of PVCS. The QRS complex of the original signal is narrow and sharp, which does not conform to the morphology of ventricular arrhythmia at all.\\
\indent 3.2 The direction of the T wave is opposite to the main QRS wave: In "reconstruct V", the T wave is in the opposite direction to the main QRS wave (if the QRS is positive, the T wave is negative), which is the manifestation of abnormal repolarization after ventricular beat. The T wave direction of the original signal is consistent with the main QRS wave, which is consistent with normal physiology.\\
\indent 3.3 Uncorrelated P waves: PVCS usually have no associated P wave or P wave is buried in QRS or T wave. There is indeed a lack of clear P waves in "reconstruct V", but the P waves are clearly discernible in the original signal.\\
\indent 3.4 Abnormal relationship between ST segment and baseline: The ST segment in "reconstruct V" is significantly elevated or depressed, suggesting myocardial ischemia or injury, while the original signal ST segment is straight and without offset.\\
Class V represents ectopic beats originating from the ventricle, and its core features are a wide and malformed QRS complex, no associated P wave, and T wave inversion. These "forcibly introduced" severe morphological anomalies are completely absent in the original signal, so class V can be clearly excluded.\\
    \end{minipage}
\end{tcolorbox}
\caption{Diagnostic report generated by MLLM.}
\label{MLLM_output}
\end{figure*}
\clearpage}

We demonstrate several specific examples of our explanation experiments in Figure~\ref{explain reconstruct}, we present three original signals~(labelled in N, S, V respectively as the features of these three labels are the most apparent and representative) and reconstruct them in signal N, S, V.
It's clear that, when the label of original signal is N, the reconstruct N signal shown in Figure~\ref{reconstruct_a} almost coincides with the original signal, however, the reconstruct S and V signal shown in Figure~\ref{reconstruct_b} and Figure~\ref{reconstruct_c} totally deviates from the original signal. It's not only a basis for classification, but also shows what features the model extracts and how the model understands the exact disease to employ the classification:

For the normal ECG signals, we can clearly distinguish P-wave, Q-wave, R-wave and S-wave(like shown in Figure~\ref{reconstruct_a}), and all the waves are normal in height and duration. And for supraventricular premature beat signals~(labelled S), the format of the P-wave is abnormal and the QRS-waves always appear earlier than normal like Figure~\ref{reconstruct_b} indicates. For Ventricular premature beat~(labelled V), the wide QRS-waves and invisible P-wave shown in Figure~\ref{reconstruct_c} are also consistent with this pathological features. Thus, these experiments phenomenons demonstrate an ablity of knowledge explanation and thus, each capsule can be seen as the information collection of different labels and the length of it is how much the information is concluded in the signal.

\subsubsection{Diagnostic Report Generation}In order to further improve the transparency and clinical utility of the model decision process, we introduce a Multi-modal Large Language Model~(MLLM, we choose Qwen3-Max here) as the final explanation generator. This module aims to convert the internal interpretable outputs of the model (i.e., reconstructed signals of different classes) into natural language diagnostic reports that can be easily understood by human doctors.

As shown in the Figure~\ref{MLLM_input} and Figure~\ref{MLLM_output}, we construct an end-to-end explanation generation pipeline: First, the model classifies the input ECG signals and generates the reconstructed waveforms corresponding to each class; Then, these original signals and multiple reconstructed signals are provided as visual input to the pre-trained MLLM. Finally, according to the preset text prompt, MLLM analyzes the visual information and automatically generates a structured report containing diagnosis conclusions and detailed reasoning basis. Because of space limitation, we only show diagnostic reports for three classes of N, P and V.

This new part not only verifies the effectiveness of the model in feature extraction and reconstruction, but more importantly, it realizes the leap from "black box" prediction to "white box" interpretation, providing clinicians with a new auxiliary diagnostic tool based on the internal mechanism of AI, which greatly enhances the credibility and operability of the model results. Our actual results, shown in Figure~\ref{MLLM_output}, demonstrate how the system can transform complex waveform differences into clear, professional medical text.

\section{Conclusion}

In this work, we have proposed an effective-while-explainable solution to ECG classification by introducing MambaCapsule, which emphasizes the capability of multi-scale Mamba in feature extracting and the interpretability of the form of Capsule network. By applying evaluation and reconstruction, the result of our model reveals great advantages in both classification score and mechanism explainablity. However, it's necessary to realize the limitation of our model, a the long training time and big memory occupied. In our future work, we aim to address these challenges by applying pre-trained parameters and optimize routing structure in Capsule network. Meanwhile, current experiments use single or averaged multi-lead signals, ignoring 12-lead spatial correlations; future work should model multi-lead jointly for more clinical alignment. As an automatic heart rhythm abnormality detection system, our model can offer doctors and patients accurate and reliable results, which have a wide range of application prospects.

\bibliographystyle{IEEEtran}
\bibliography{name.bib}


\newpage

\section{Biography Section}
\vspace{-33pt}
\begin{IEEEbiography}
[{\includegraphics[width=1in,height=1.25in,clip,keepaspectratio]{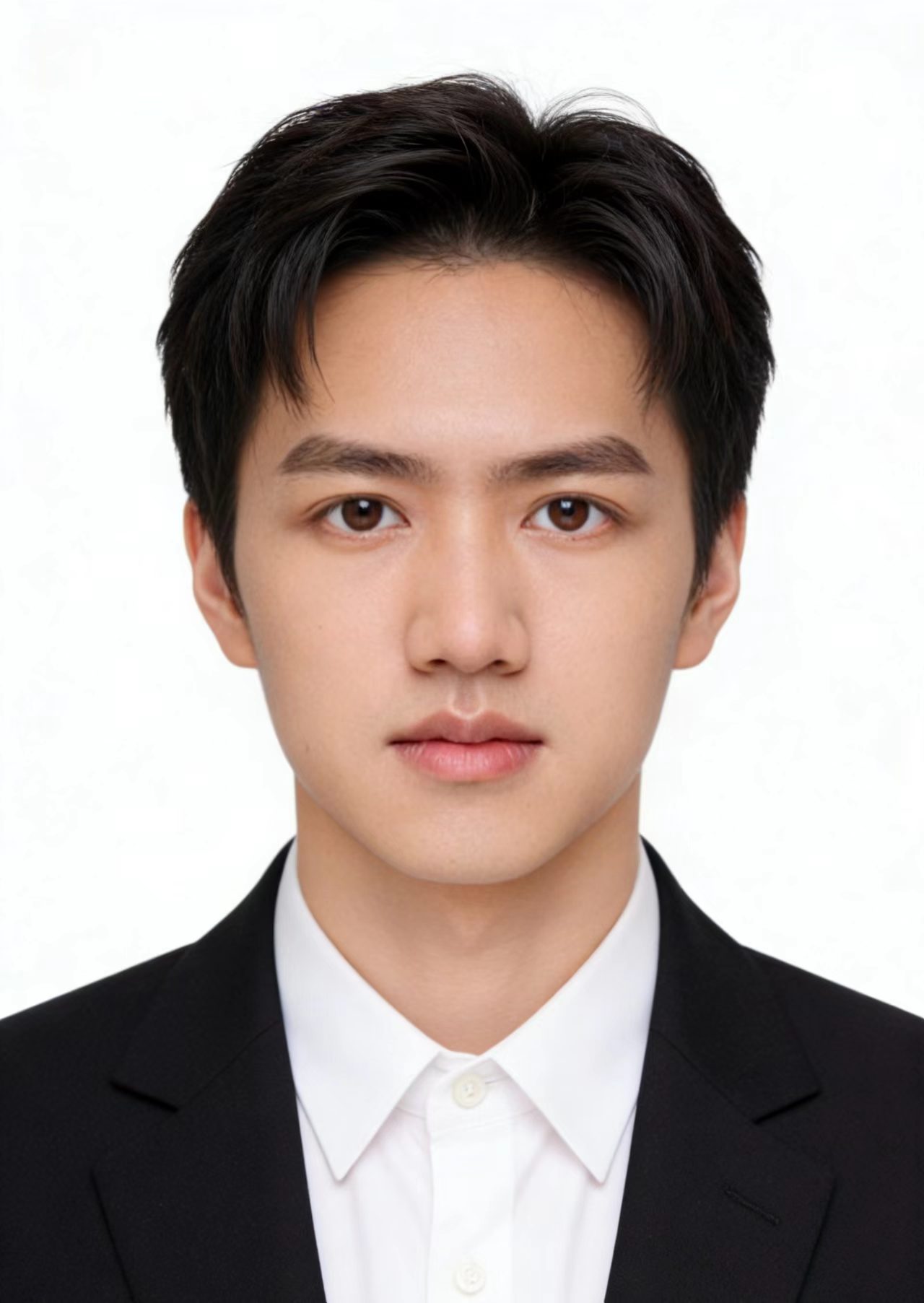}}]{Yinlong Xu} received the B.S. degree from Beihang University, Beijing, China. He is currently pursuing the M.S. degree in the major of artificial intelligence in Zhejiang University, Hangzhou, China. His research interests include large language models, reasoning paradigms, and medical artificial intelligence.
\end{IEEEbiography}
\vspace{-33pt}
\begin{IEEEbiography}
[{\includegraphics[width=1in,height=1.25in,clip,keepaspectratio]{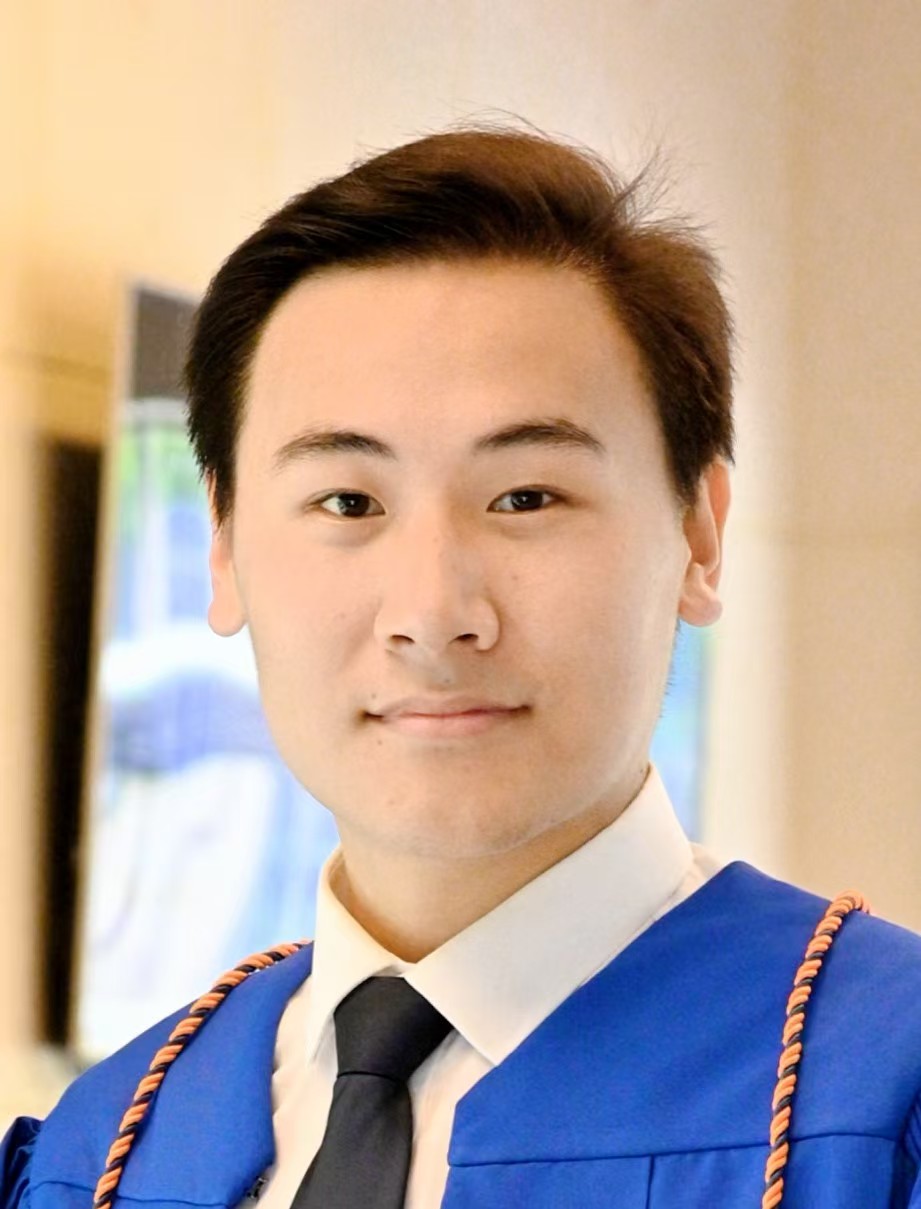}}]{Zitai Kong}
 received the B.S. degree in ECE from Zhejiang University, Hangzhou, China. He is currently pursuing the Ph.D. degree in the College of Computer Science and Technology, Zhejiang University, Hangzhou, China. His research interests include deep generative models, espacially on AI for science.
\end{IEEEbiography}
\vspace{-33pt}
\begin{IEEEbiography}[{\includegraphics[width=1in,height=1.25in,clip,keepaspectratio]{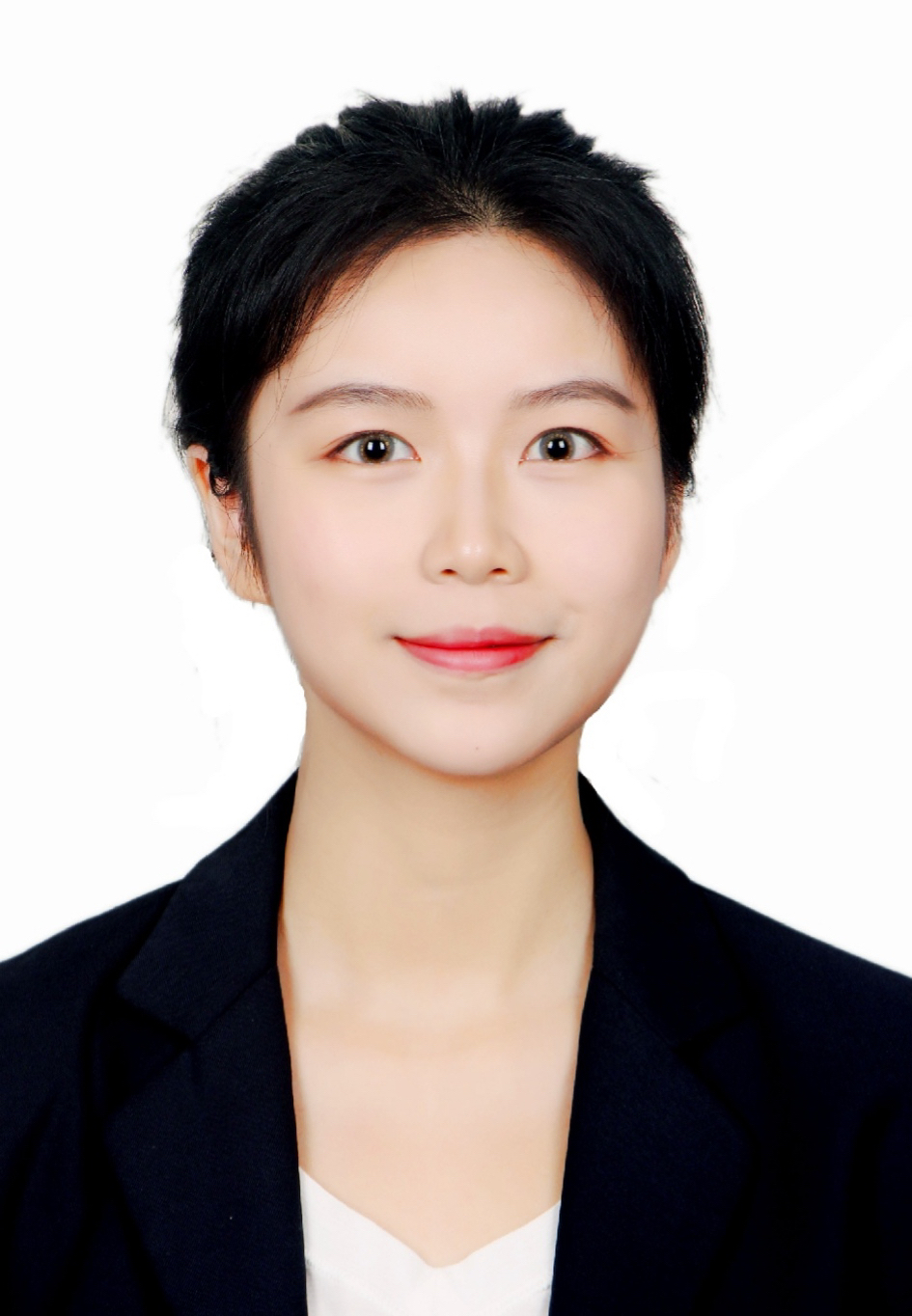}}]{Yixuan~Wu} 
received the B.S. degree from the Zhejiang University, Hangzhou, China. She is currently working toward the Ph.D. degree in the major of big data and health science in Zhejiang University, Hangzhou, China. Her research interests includes deep learning, machine learning, especially on the multimodal learning and medical intelligence.
\end{IEEEbiography}
\vspace{-33pt}
\begin{IEEEbiography}
[{\includegraphics[width=1in,height=1.25in,clip,keepaspectratio]{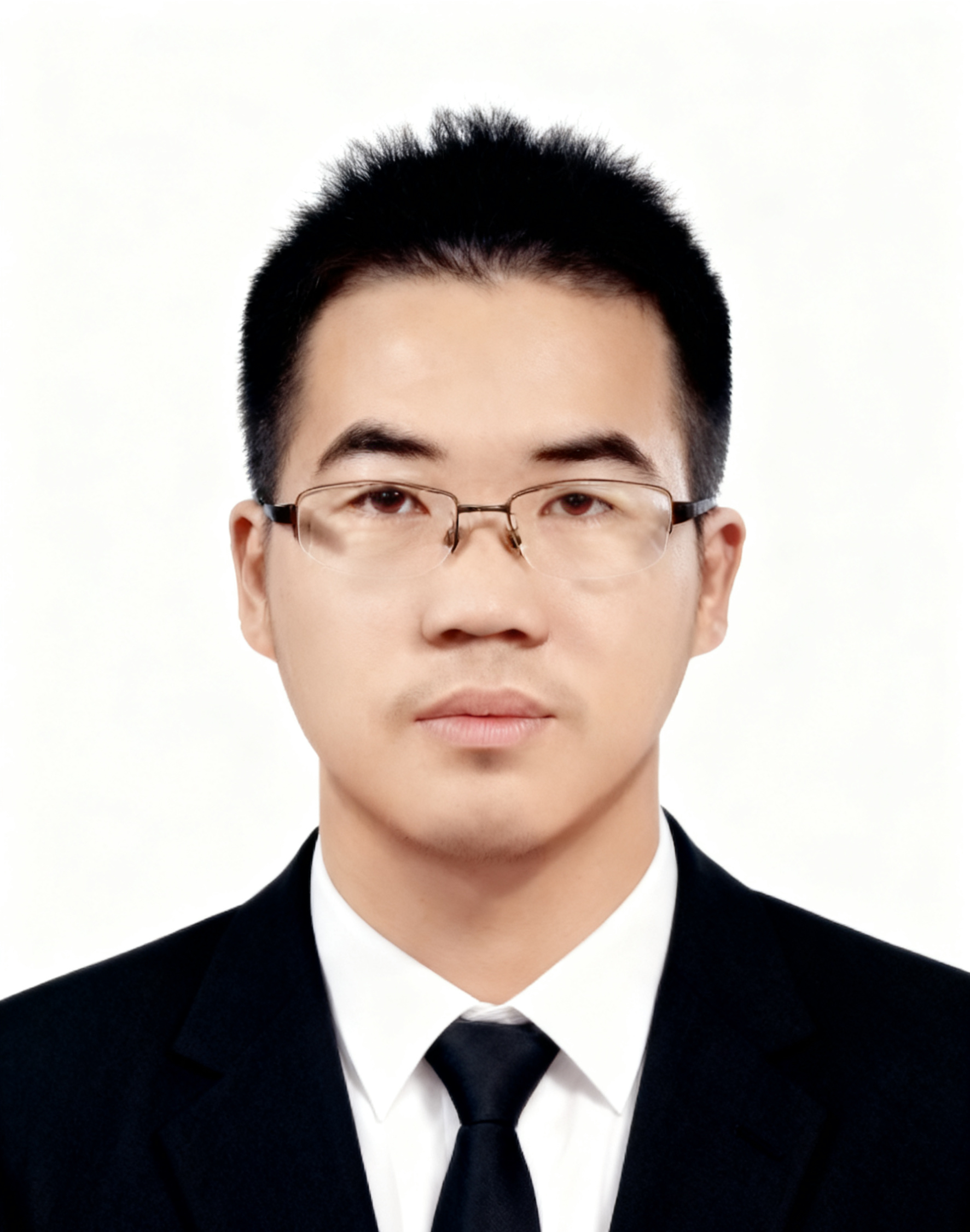}}]{Yue Wang}
 is currently working toward the Ph.D. degree in Computer Science and Technology at Zhejiang University, China. His research interests include signal representation in medical applications and multimodal embodied intelligence for medicine.
\end{IEEEbiography}
\vspace{-33pt}
\begin{IEEEbiography}
[{\includegraphics[width=1in,height=1.25in,clip,keepaspectratio]{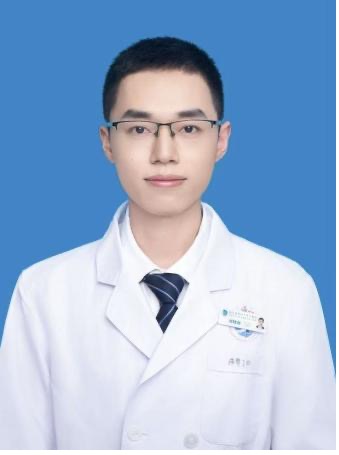}}]{Xiaoqiang Liu} received the master’s degree in Gastroenterology from Fujian Medical University, in 2016, where his research focused on gastrointestinal diseases and clinical digestive endoscopy. He is currently affiliated with Quanzhou First Hospital, where his research interests include digestive system disorders and endoscopic diagnosis and treatment. His recent project involves the development of intelligent diagnostic systems for gastrointestinal disease analysis.
\end{IEEEbiography}
\vspace{-33pt}
\begin{IEEEbiography}
[{\includegraphics[width=1in,height=1.25in,clip,keepaspectratio]{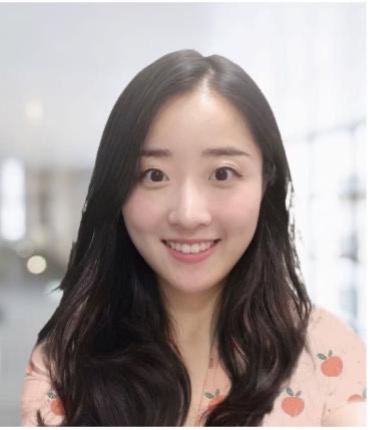}}]{Yingzhou Lu} is a postdoctoral researcher at Stan-ford University, she obtained her Ph.D. in Artificial Intelligence and Computational Biology from Virginia Tech. With eight years of expertise in machine learning and genomics,she has cultivated a deep understanding of genomics data analysis, graph theory and multi-omics data integration. Her research focuses on employing advanced computational techniques to analyze genomics datasets, aiming to shed light on disease development and progression.
\end{IEEEbiography}
\vspace{-33pt}
\begin{IEEEbiography}
[{\includegraphics[width=1in,height=1.25in,clip,keepaspectratio]{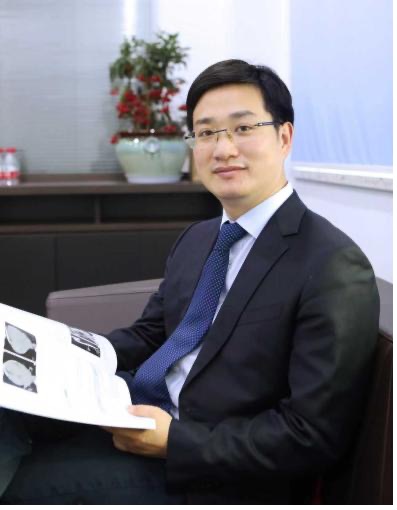}}]{Jian Wu} is a Qiushi Distinguished Professor at Zhejiang University. He serves as the Director of the Provincial Key Laboratory of Artificial Intelligence for Medical Imaging, and an expert member of the expert panel of the national standardization technical committee responsible for AI-based medical devices. His research focuses on medical artificial intelligence. In recent years, he has led seven projects funded by the National Natural Science Foundation of China, undertaken four sub-projects under China’s National Key R\&D Program, and directed two major industry-sponsored projects. He has published over 200 SCI/EI-indexed papers as first or corresponding author, with more than 19,000 citations in total; his most-cited single paper has received over 3,700 citations.
\end{IEEEbiography}
\vspace{-33pt}
\begin{IEEEbiography}
[{\includegraphics[width=1in,height=1.25in,clip,keepaspectratio]{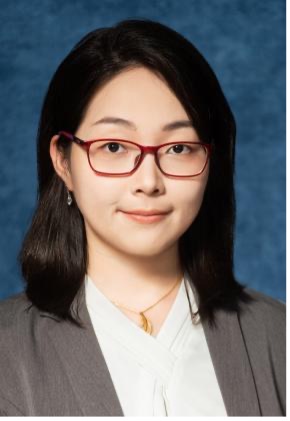}}]{Hongxia Xu} received the Ph.D. degree in School of Chemical Engineering and Bioengineering, Zhejiang University. From 2019 to 2023, she worked as a postdoctoral researcher at Zhejiang University, conducting research on the development of polymer materials for tumor immunotherapy as well as medical artificial intelligence. From 2023 to now, she worked as a researcher and focusing on medical large language models. In recent years, she has published over 80 SCI and conference papers in international journals and top computer science conferences such as Biomaterials, Nano Letters, NeurIPS, and ICLR. She has been awarded the Second Prize of the Science and Technology Award of the Chinese Institute of Electronics(2023), the Second Prize for Innovation Achievement in China Industry-University-Research Institute Collaboration Association(2023), and the Innovation Award of the Invention and Entrepreneurship Award, China Association of Inventions(2025).
\end{IEEEbiography}


\vfill

\end{document}